\pdfoutput=1

\documentclass[11pt]{article}
\usepackage{setspace}
\usepackage[]{acl}

\usepackage{times}
\usepackage{amsmath,amsthm,amsfonts,amssymb,bm}
\usepackage{latexsym}
\usepackage{hyperref}
\usepackage{inconsolata}
\usepackage{hyperref}
\usepackage{url}
\usepackage{xcolor}
\usepackage{adjustbox} 
\usepackage{listings}
\usepackage{multirow}
\usepackage{subcaption}
\usepackage{graphicx}
\usepackage{float}
\usepackage{enumitem}
\usepackage{caption}
\usepackage{tabularx}
\usepackage{adjustbox}
\newcolumntype{C}{>{\centering\arraybackslash}X}
\usepackage{array} 

\usepackage{booktabs} 
\usepackage[T1]{fontenc}

\usepackage[utf8]{inputenc}

\usepackage{microtype}

\usepackage{inconsolata}

\usepackage[]{todonotes} 

\newtheorem{theorem}{Theorem}

\newcommand{\codefont}{\fontfamily{lmtt}\selectfont}

\usepackage{soul}
\definecolor{aigold}{RGB}{244,210, 1} 
\definecolor{aigreen}{RGB}{245, 255, 249}

\definecolor{humanpurple}{RGB}{235, 222, 240} 

\definecolor{commentgray}{RGB}{86, 101, 115}

\definecolor{aired}{RGB}{255,180,181}

\usepackage{listings}
\usepackage{parcolumns}
\lstdefinestyle{datalogstyle}{
	basicstyle={\codefont \small},  
	xleftmargin={6pt},
        xrightmargin={6pt},
        breakindent=0pt,
	frame=tb,
	stepnumber=1,
	firstnumber=1,
	numberfirstline=true,
	tabsize=2,
	showtabs=false,
	showspaces=false,
	showstringspaces=false,
	extendedchars=true,
	breaklines=true,
	columns=fullflexible,
	keepspaces=true,
	escapeinside={@}{@},
	firstnumber=last,
	captionpos=b,
	commentstyle=\color{black!65},
	numberstyle=\tiny\color{black!65},
	stringstyle=\color{codepurple},
	breakatwhitespace=false, 
	keepspaces=true,                 
	numbersep=5pt,                  
	showspaces=false,                
	showstringspaces=false,
	showtabs=false,
	aboveskip={0.8\baselineskip},
	belowskip={0.2\baselineskip},
	backgroundcolor=\color{aigreen},
}
\title{CoCA: Fusing Position Embedding with Collinear Constrained Attention in Transformers for Long Context Window Extending}

\author{
    Shiyi Zhu, Jing Ye, Wei Jiang, Siqiao Xue, Qi Zhang, Yifan Wu, Jianguo Li \\
    Ant Group \\
    \texttt{\{zhushiyi.zsy, qianye.yj, shouzhi.jw, lijg.zero\}@antgroup.com}
}

\begin{document}

\maketitle
\begin{abstract}
Self-attention and position embedding are two key modules in transformer-based Large Language Models (LLMs). However, the potential relationship between them is far from well studied, especially for long context window extending. In fact, anomalous behaviors harming long context extrapolation exist between Rotary Position Embedding (RoPE) and vanilla self-attention unveiled by our work.
To address this issue, we propose a novel attention mechanism, CoCA (\textbf{Co}llinear \textbf{C}onstrained \textbf{A}ttention). Specifically, we enforce a collinear constraint between $Q$ and $K$ to seamlessly integrate RoPE and self-attention.
While only adding minimal computational and spatial complexity, this integration significantly enhances long context window extrapolation ability. We provide an optimized implementation, making it a drop-in replacement for any existing transformer-based models.
Extensive experiments show that CoCA performs extraordinarily well in extending context windows. A CoCA-based GPT model, trained with a context length of 512, can seamlessly extend the context window up to 32K (60$\times$), without any fine-tuning.
Additionally, by dropping CoCA in LLaMA-7B, we achieve extrapolation up to 32K within only 2K training length.
Our code is publicly available at: 
\href{https://github.com/codefuse-ai/Collinear-Constrained-Attention}{https://github.com/codefuse-ai/Collinear-Constrained-Attention}

\end{abstract}

\section{Introduction}

\begin{figure}[ht!]
    \centering
    \includegraphics[width=0.9\linewidth]{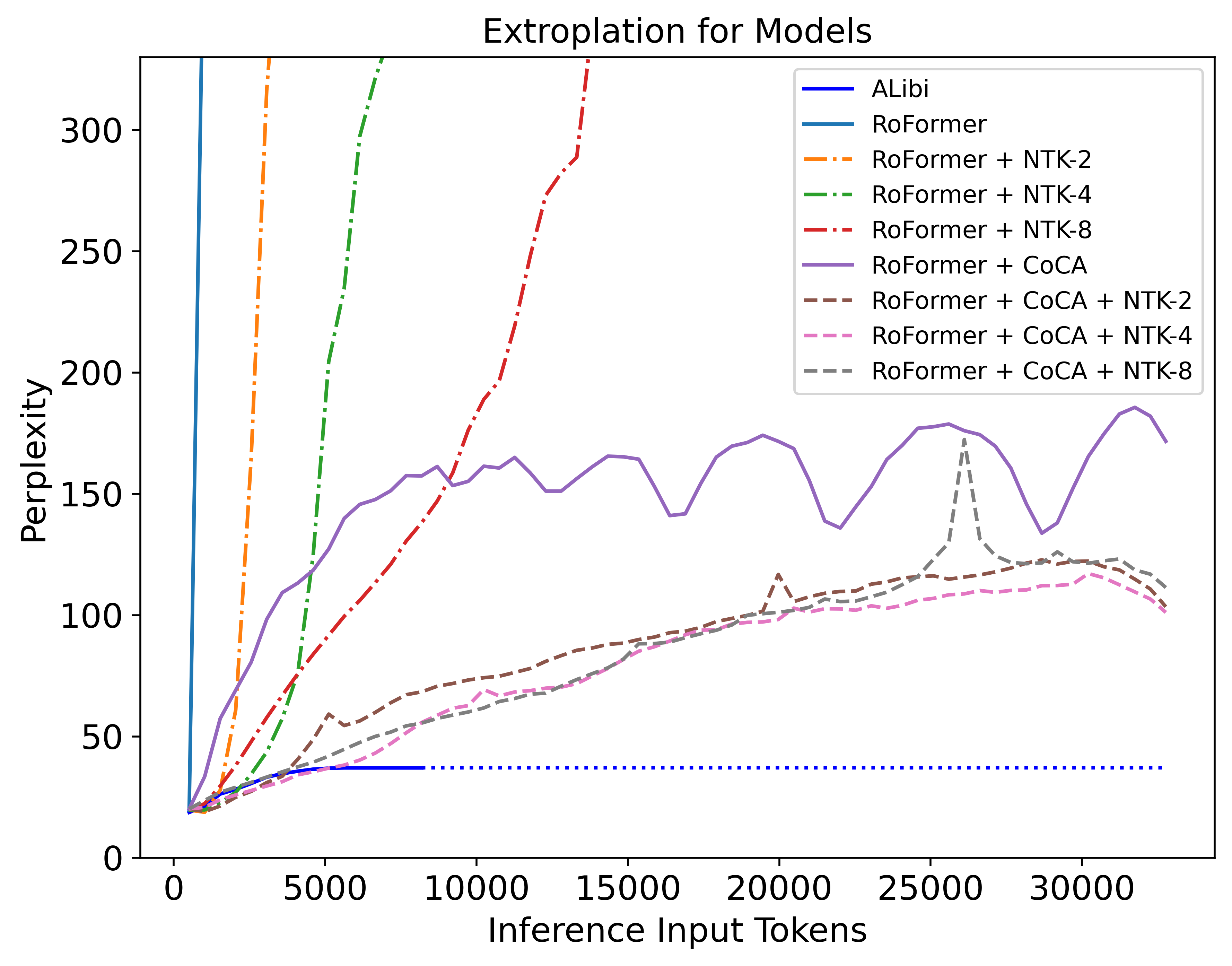}
    \caption{Perplexity evaluation on 100 PG-19 documents with a sliding window strategy (Stride = 512). The perplexity of RoFormer \citep{Su2021RoFormerET} sharply exceeds 1000 beyond its training length, while CoCA maintains a low plateau even at 60 $\times$ its training length. ALibi \citep{Press2021TrainST} encounters Out of Memory (OOM) issues for input \(N_{max}>\) 8000 due to flash-attention \citep{dao2022flashattention} incompatibility, we suppose it maintains perplexity for \(N_{max}>\) 8000.}
    \label{fig:ppl}
\end{figure}

In the seminal work of Transformer \citep{vaswani2017attention}, it claims the ability of "extrapolating to sequence length longer than the ones encountered during training". This is an ideal hypothesis, but actually not work in practice for vanilla Transformer. Several subsequent works, collectively known as long context extrapolation, have delved into exploring the capabilities of large language models (LLMs) trained within the range of $[1, N-1]$ to effectively extend the testing sequence $\geq N$.
 
Existing studies primarily focus on attention kernel \citep{beltagy2020longformer,ding2023longnet, han2023lm} or position embedding \citep{huang2023advancing}, often neglecting the intrinsic relationship between the two key modules. 
Attention bias is an alternative to the explicit encoding of positional information. ALibi \citep{Press2021TrainST} and KERPLE \citep{Chi2022KERPLEKR}, incorporate heuristic and compositional triangle kernel-based negative causal attention bias, respectively. While these approaches effectively manage to maintain low perplexity, they fall short in capturing long-range dependencies due to introducing local hypotheses to context tokens. 
Another branch of methods involve simply scaling Rotary Position Embedding (RoPE) \citep{Su2021RoFormerET} to extrapolate the inference context length with minimal or no fine-tuning. For instance, Position Interpolation (PI) \citep{Chen2023ExtendingCW} employs linear scaling on each position number from \(n\) to \(n/k\), where $k$ is the extrapolation ratio. NTK-aware Scaled RoPE \citep{NTK} and Dynamic-NTK \citep{dynamicNTK} combine high-frequency extrapolation and low-frequency interpolation. They scale the basis in RoPE upon the sequence length to adapt to the unseen position indices. However, these methods primarily alleviate the problem of modeling the rotation angles in out-of-distribution positions, without recognizing the intrinsic correlation between attention matrices and rotation angles. Therefore, these methods still suffer from a limited context window extending ratio.          

Here, we present a new perspective on the relationship between position embedding (with a focus on RoPE) and the self-attention mechanism. In a nutshell,  RoPE utilizes a rotation matrix to encode absolute positions while simultaneously incorporating explicit relative position dependencies within the self-attention formulation \citep{Su2021RoFormerET}. It is designed based on the relative angular difference between the queries ($Q$) and keys ($K$). However, latent relationships exist between $Q$ and $K$, as these two matrices are directly multiplied. We demonstrate that incorrect initialization of the angle between $Q$ and $K$ in RoPE leads to undesirable behavior around the context window boundary, harming its performance for context extrapolation. 

\begin{figure*}[htp]
    \centering
    \includegraphics[width=1\linewidth]{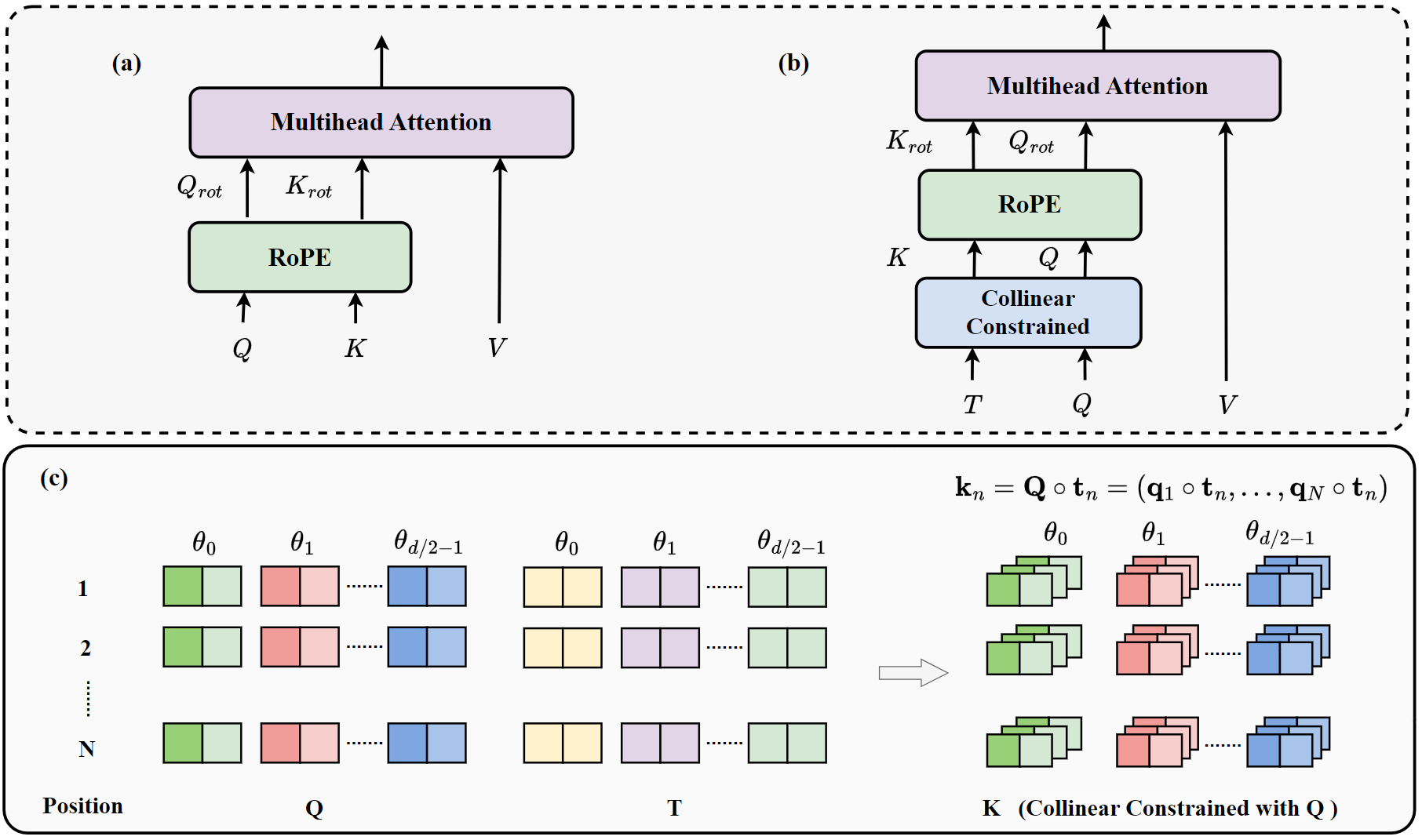}
    \caption{Architecture comparison between RoFormer and CoCA.
    (a) RoFormer; (b) CoCA; 
    (c) The implementation detial of \textit{K} in CoCA. \textit{Q}, \textit{T}, and \textit{V} are produced using projection matrices identical to those employed in the vanilla self-attention. \textit{T} undergoes a halving operation, with the other half being duplicated. \textit{K} is then computed as the element-wise product of \textit{Q} and \textit{T}, adhering to a collinear constraint with \textit{Q}. Note that $\mathbf{k}_n \in \mathbb{R}^{N \times d}$, where $n\in [1,N]$ is the positional index of key, $d$ is the head dimension, $N$ is the sequence length.}
    \label{fig:model}
\end{figure*}

To address this undesirable behavior
, we propose an innovative architecture called Collinear Constrained Attention (CoCA). Specifically, we enforce a collinear constraint between $Q$ and $K$ by initializing the angle between every two hidden dimensions in the $Q$ and $K$ vectors to 0. This allows for a seamless integration of RoPE and self-attention.
The model architecture and comparison with RoFomer \cite{Su2021RoFormerET} is illustrated in Figure \ref{fig:model}. 

Extensive experiments show that a CoCA-based GPT model, trained within 512 context length, seamlessly extends the context window up to 32K (60x) without perplexity divergence. A comprehensive comparison between our method and existing methods is presented in Figure \ref{fig:ppl}. Furthermore, it enhances long-context retrieval ability, achieving a passkey retrieval accuracy of 50\%+ even when extrapolating to 16x longer than its training context length by applying Dynamic-NTK \citep{dynamicNTK}. Additionally, by dropping CoCA in LLaMA-7B, we achieve extrapolation up to 32K within only 2K training length.

Our main contributions can be summarized as follows:
\begin{itemize}[itemsep= 0.1pt,topsep = 0.1pt,partopsep=0.1pt]
\item We unveil undesirable context boundary behavior resulting from the absence of modeling the relationship between position embeddings and self-attention. 
\item To tackle the undesirable context boundary behavior, we propose Collinear Constrained Attention (CoCA) to seamlessly integrate the position embeddings and self-attention, achieving excellent long context window extrapolation performance. 
\item CoCA extends its context window from 512 to 32K without fine-tuning, achieving over 50\% accuracy even when 16 $\times$ longer than its training length. Using CoCA in LLaMA-7B, we achieve extrapolation up to 32K within just 2K training length.
\item CoCA introduces minimal computational and spatial complexity compared to vanilla self-attention. We provide an optimized implementation of CoCA, making it able to be a seamless drop-in replacement for existing transformer-based models.
\end{itemize}

\section{Method}
In this section, we describe our proposed \textbf{Co}llinear \textbf{C}onstrained \textbf{A}ttention (CoCA). We begin with introducing the background theory of RoPE \cite{Su2021RoFormerET} in Section \ref{Rotary Position Embedding}, and then analyze the anomalous behaviors between the attention matrices and RoPE in Section \ref{anomalous analysis}. Finally, we introduce the proposed method CoCA in section \ref{Collinear Constrained Attention} and derive a slack constraint version of CoCA in Section \ref{Slacking the constraint on query}, respectively.

\subsection{Rotary Position Embedding}
\label{Rotary Position Embedding}
Position embedding is a crucial component in transformer-based models. Here we focus on Rotary Position Embedding (RoPE) \citep{Su2021RoFormerET}, which is widely used by LLMs including LLaMA \citep{Touvron2023LLaMAOA}, LLaMA-2 \citep{Touvron2023Llama2O}, GPT-NeoX \citep{black-etal-2022-gpt} and Qwen \cite{qwen}. Suppose the positional index is an integer $n\in[1,N]$, and the corresponding input vector $\mathbf{x} = [x_0,x_1,...,x_{d-1}]^{\text{T}}$, where \(N\) is the sequence length, $d$ is the dimension of the attention head. RoPE defines a vector-valued complex function $f(\mathbf{x},n)$:
\begin{equation}
\small
\begin{aligned}
f(\mathbf{x},n) &= [(x_0+ix_1)e^{in\theta_0},(x_2+ix_3)e^{in\theta_1}, \\
          &\quad \ldots, (x_{d-1}+ix_d)e^{in\theta_{d/2-1}}]^{\text{T}},  \\
&where \ \ \theta_{j}=\text{B}^{-2j/d},
\end{aligned}
\label{eq:1}
\end{equation}
in this paper, the base $\text{B}=10,000$.

After the application of RoPE, the transformed vectors for query ($\mathbf{q}$) and key ($\mathbf{k}$) become $f(\mathbf{q}, m)$ and $f(\mathbf{k}, n)$, respectively. Here, $m,n \in [0,N]$ represent the positional indices of $\mathbf{q}$ and $\mathbf{k}$. The attention operation is computed as the dot product between $f(\mathbf{q}, m)$ and $f(\mathbf{k}, n)$, defined as follows:
\begin{equation}
\small
\begin{aligned}
&a(m,n) =\text{Re}(\langle f(\mathbf{q},m),f(\mathbf{k},n)\rangle)\\
       &=\text{Re}\left[\sum_{j=0}^{d/2-1}(q_{2j}+iq_{2j+1})(k_{2j}-ik_{2j+1})e^{i(m-n)\theta_j} \right] \\
       &=\sum_{j=0}^{d/2-1}[(q_{2j}k_{2j}+q_{2j+1}k_{2j+1})\cos((m-n)\theta_j) \\
       &\quad +(q_{2j}k_{2j+1}-q_{2j+1}k_{2j})\sin((m-n)\theta_j)] \\
\end{aligned}
\label{eq:2}
\end{equation}
The attention score $a(m-n)$ depends on the relative position $(m - n)$. 

\subsection{Anomalous Behavior between RoPE and Attention Matrices}
\label{anomalous analysis}
\begin{figure*}[htp]
\centering
\includegraphics[width=\linewidth]{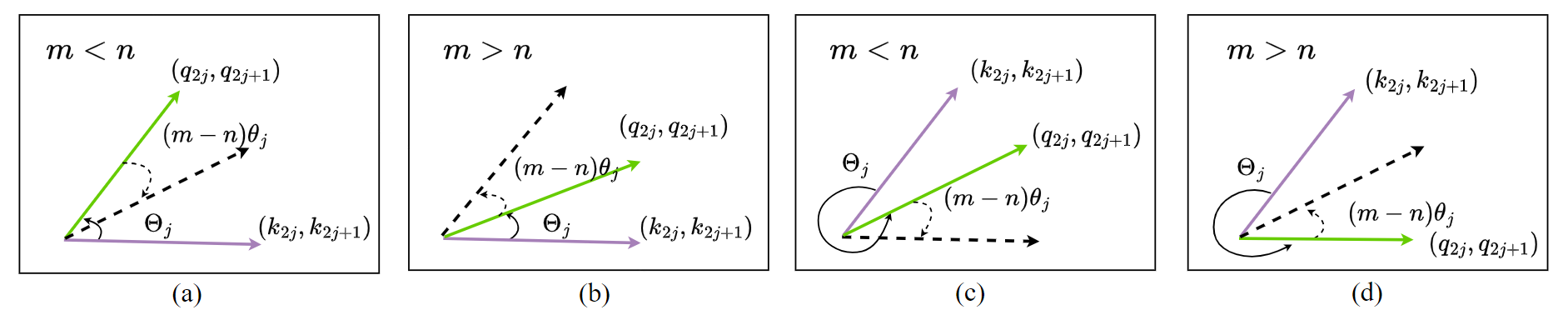}
\caption{Anomalous behavior of RoPE in 2-D plane. The inner product of vectors $\mathbf{q}_j$ and $\mathbf{k}_j$ is contingent upon the relative angle $\theta{(\mathbf{q}_j,\mathbf{k}_j)}$, defined as $\Theta_j + (m-n)\theta_j$. Here, $\Theta_j$ represents the initial angle, and $(m-n)\theta_j$ signifies the position-dependent rotation angle. (a) $m < n$ and $\Theta_j \leq \pi$. (b) $m > n$ and $\Theta_j \leq \pi$. (c) $m < n$ and $\Theta_j > \pi$. (d) $m > n$ and $\Theta_j > \pi$.}
\label{fig:Anomalous Behavior Analysis}
\end{figure*}

After applying RoPE, the attention score $a(m-n)$ can be interpreted as the sum of $d/2$ inner products of complex numbers, as illustrated in Equation (\ref{eq:2}). For any pair of $\mathbf{q}_j=(q_{2j},q_{2j+1})$ and $\mathbf{k}_j=(k_{2j},k_{2j+1})$, which is the 2-dimensional slicing of $\mathbf{q}$ (or $\mathbf{q}_m$) and $\mathbf{k}$ (or $\mathbf{k}_n$), we introduce the initial angle $\Theta_j$ between them, measured counterclockwise from $\mathbf{k}_j$ to $\mathbf{q}_j$ in the complex plane. Throughout our analysis, we keep the position of $\mathbf{k}_j$ fixed, systematically rotating $\mathbf{q}_j$ to comprehensively examine their relative positions. The final angle between $\mathbf{q}_j$ and $\mathbf{k}_j$ is represented as $\theta{(\mathbf{q}_j,\mathbf{k}_j)} = \Theta_j + (m-n)\theta_j$, where $m$ and $n$ are positional indices of $\mathbf{q}_j$ and $\mathbf{k}_j$.

In this concept, the attention score can be formulized as:
\begin{equation}
\small
\begin{aligned}
&a(m,n) =\sum_{j=0}^{d/2-1}|\mathbf{q}_j||\mathbf{k}_j|\cos(\theta{(\mathbf{q}_j,\mathbf{k}_j)}) 
\end{aligned}
\label{eq:inner product}
\end{equation}

Refer to Figure \ref{fig:Anomalous Behavior Analysis} for a visual representation of this concept for any individual $j\in [0,d/2]$ in the 2-D subspace. There are four distinct scenarios between $\mathbf{q}_j$ and $\mathbf{k}_j$ after rotation.

(1) \textbf{Scenario (b) and (c):} When $m > n$ and $\Theta_j \leq \pi$, or $m < n$ and $\Theta_j > \pi$, the value of $\cos(\theta{(\mathbf{q}_j,\mathbf{k}_j)})$ between $\mathbf{q}_j$ and $\mathbf{k}_j$ decreases with the expanding distance between $m$ and $n$. In these 2 scenarios, no anomalous behavior is observed, as the attention score naturally decreases with the positional distance. This trend persists until the relative angle $\theta{(\mathbf{q}_j,\mathbf{k}_j)}$ rotates beyond the boundary of $\pi$. 

(2) \textbf{Scenario (a) and (d):} When $m < n$ and $\Theta_j \leq \pi$, or $m > n$ and $\Theta_j > \pi$, intriguing phenomena emerge. As the distance between $m$ and $n$ grows, the value of $\cos(\theta{(\mathbf{q}_j,\mathbf{k}_j)})$ between $\mathbf{q}_j$ and $\mathbf{k}_j$ paradoxically increases. This anomaly has a notable impact on attention scores, particularly affecting the $\tau$ closest tokens. In this context, $\tau$ is defined as $\Theta_j/\theta_j$ for scenario (a) and $(2\pi-\Theta_j)/\theta_j$ for scenario (d). Consequently, attention scores for these tokens are abnormally diminished.

For bidirectional language models, all four cases may occur. For causal models, only scenario (b) and (d) manifest, as $m$ consistently exceeds $n$. 

The attention score $a(m-n)$ is the sum of $d/2$ inner-products, one of them turns out anomalous may be insignificant, however, experiments confirmed this significance. Further analysis of this rotary borders anomalous behaviour is discussed in Appendix \ref{Rotary Borders Analysis}.

\subsection{Collinear Constrained Attention}
\label{Collinear Constrained Attention}
 
To tackle the anomalous behavior between RoPE and attention matrices, we propose a novel approach called \textbf{Co}llinear \textbf{C}onstrained \textbf{A}ttention (CoCA). Specifically, by applying a collinear constraint to any pair of $\mathbf{q}_j=(q_{2j},q_{2j+1})$ and $\mathbf{k}_j=(k_{2j},k_{2j+1})$, we seamlessly integrate RoPE into self-attention mechanism, achieving long context extrapolation.

To formalize this, considering a sequence of $N$ input tokens $\mathbb{S}_N=\{w_n\}_{n=1}^N$, with corresponding word embeddings $\mathbb{E}_N=\{\mathbf{x}_n\}_{n=1}^N$, where $\mathbf{x}_n\ \in \mathbb{R}^d$ is the $d$-dimensional word embedding vector of token $w_n$ without position information. First, the queries $\mathbf{q}_m$ are obtained:
\begin{equation}
\small 
\begin{aligned}
\mathbf{q}_m={\textbf{W}}_Q\mathbf{x}_m, \forall m\in [1,N]
\end{aligned}
\label{eq:21}
\end{equation}

Next, we derive the keys $\mathbf{k}_n$ with collinear constraints. This begins with the introducing of the constraint coefficient $\mathbf{t}_n$ for each token position $n$, as depicted in Equation (\ref{eq:22}).

\begin{equation}
\small 
\begin{aligned}
\mathbf{t}_n={\textbf{W}}_T\mathbf{x}_n, \forall n\in [1,N]
\end{aligned}
\label{eq:22}
\end{equation}

Next, Equation (\ref{eq:5}) imposes the collinearity condition on the coefficients ${t}_{2j}$ and ${t}_{2j+1}$, where $\mathbf{t}_n=[t_0,t_1,...,t_{d-1}]^{\text{T}}$, ensuring that each pair is identical. This step effectively duplicates each 2-dimensional segment of the tensor.
\begin{equation}
\small 
\begin{aligned}
&{t}_{2j} = {t}_{2j+1} , \forall j \in [0, d/2-1] \\
&\mathbf{t}_n=\text{Relu}(\mathbf{t}_n)
\end{aligned}
\label{eq:5}
\end{equation}

Subsequently, the keys are calculated as shown in Equation (\ref{eq:6}), where $\mathbf{k}_n$ are represented by the element-wise multiplication of $\mathbf{Q}=(\mathbf{q}_1,...,\mathbf{q}_N)$ and $\mathbf{t}_n$. This results in an expansion of dimensionality, as $\mathbf{k}_n \in \mathbb{R}^{N \times d}$ now includes an additional sequence length dimension. We address potential memory pressure by optimizing tensor contractions, ensuring no net increase in memory consumption. 
For an in-depth analysis, please refer to Appendix \ref{Computational and spatial complexity}.
\begin{equation}
\small 
\begin{aligned}
&\mathbf{k}_n=\mathbf{Q}\odot \mathbf{t}_n =(\mathbf{q}_1\circ \mathbf{t}_n,...,\mathbf{q}_N\circ \mathbf{t}_n)
\label{eq:6}
\end{aligned}
\end{equation}
After that, we apply RoPE on $Q$ and $K$, with the function $f$ detailed in Equation (\ref{eq:1}).
\begin{equation}
\small 
\begin{aligned}
&f(\mathbf{q}_m)=f(\mathbf{q}_m,m) \\
&f(\mathbf{k}_n)=f(\mathbf{Q}\odot \mathbf{t}_n,n) = f(\mathbf{Q},n) \odot \mathbf{t}_n
\label{eq:7}
\end{aligned}
\end{equation}
Finally, the attention score of CoCA would be:
\begin{equation}
\small
\begin{aligned}
a(m,n) &=\text{Re}(\langle f(\mathbf{q}_m,m), f(\mathbf{q}_m,n) \circ \mathbf{t}_n \rangle)
\label{eq:8}
\end{aligned}
\end{equation}
Equation (\ref{eq:8}) illustrates the additional dimension of the keys in our CoCA mechanism. Specifically, it maps the index of each query to the additional dimension, establishing a collinear relationship between the $n$-th key and the $m$-th query. This is a critical aspect of our method.

\subsection{Slacking the Constraint on Query}
\label{Slacking the constraint on query}

In Section \ref{Collinear Constrained Attention}, we present a theoretically precise solution for CoCA. However, practical implementation faces challenges due to the complexity of $O(N^2d)$ when storing $f(\mathbf{Q},n)$. To address this issue, we provide a dual implementation with $O(Nd)$ complexity in this section and prove their equivalence.

\begin{theorem}
    (Dual implementation of CoCA) For any attention score defined in Equation (\ref{eq:8}), there exists an equivalent form as follows:
    \begin{equation}
    \small 
        \begin{aligned}
        a(m,n) &=\text{\rm{Re}}(\langle f(\mathbf{q}_m,m), \mathbf{q}_m \circ f(\mathbf{t}_n,n)  \rangle)
        \label{eq:slack}
        \end{aligned}
    \end{equation}
    with constraint:
    \begin{equation}
    \small
    \begin{aligned}
    q_{2j} = q_{2j+1},\forall j\in [0,d/2-1]
    \end{aligned}
    \label{eq:query constraint}
    \end{equation}
\label{th:2.1}
\end{theorem}

\noindent \textit{\textbf{Proof}}: 
The proof consists of two steps. 

\noindent \textit{Step 1.} We prove that, by imposing the constraint $q_{2j} = q_{2j+1}$, $\forall j\in [0,d/2-1]$, $\text{\rm{Re}}(\langle f(\mathbf{q}_m,m), \mathbf{q}_m \circ f(\mathbf{t}_n,n)  \rangle)$  is equivalent to $\text{Re}(\langle f(\mathbf{q}_m,m), f(\mathbf{q}_m,n) \circ \mathbf{t}_n \rangle)$.

To see this, we calculate the difference between $ f(\mathbf{q}_m,n) \circ \mathbf{t}_n$ and $\mathbf{q}_m \circ f(\mathbf{t}_n,n)$:
\begin{equation}
\small
\begin{aligned}
&f(\mathbf{q}_m,n)\circ \mathbf{t}_n-\mathbf{q}_m\circ f(\mathbf{t}_n,n) \\
&= \left(\begin{array}{cc}
  & t_0 (q_0\cos n\theta_0-q_{1}\sin n\theta_0)  \\
  & t_1 (q_0\sin n\theta_0+q_{1}\cos n\theta_0)  \\
  & \ldots    \\
  & t_{d-2} (q_{d-2}\cos n\theta_{d/2-1}-q_{d-1}\sin n\theta_{d/2-1})  \\
  & t_{d-1} (q_{d-2}\sin n\theta_{d/2-1}+q_{d-1}\cos n\theta_{d/2-1})  \\
\end{array}\right) \\
&-
\left(\begin{array}{cc}
  & q_0 (t_0 \cos n\theta_0-t_1 \sin n\theta_0)  \\
  & q_1 (t_0 \sin n\theta_0+t_1 \cos n\theta_0)  \\
  & \ldots    \\
  & q_{d-2}(t_{d-2} \cos n\theta_{d/2-1}-t_{d-1} \sin n\theta_{d/2-1})  \\
  & q_{d-1}(t_{d-2} \sin n\theta_{d/2-1}+t_{d-1} \cos n\theta_{d/2-1})  \\
\end{array}\right) \\
\end{aligned}
\label{theorem_dual_coca_1}
\end{equation}
Recall that $t_{2j}=t_{2j+1}, \forall j\in [0,d/2-1]$ (see Equation (\ref{eq:5})), Equation (\ref{theorem_dual_coca_1}) is equivalent to:

\begin{equation}
\small
\begin{aligned}
&f(\mathbf{q}_m,n)\circ \mathbf{t}_n-\mathbf{q}_m\circ f(\mathbf{t}_n,n) \\
&=\left(\begin{array}{cc}
  & t_0 (q_0-q_{1})\sin n\theta_0  \\
  & t_1 (q_0-q_{1})\sin n\theta_0  \\
  & \ldots    \\
  & t_{d-2} (q_{d-2}-q_{d-1})\sin n\theta_{d/2-1}  \\
  & t_{d-1} (q_{d-2}-q_{d-1})\sin n\theta_{d/2-1}  \\
\end{array}\right) 
\end{aligned}
\label{theorem_dual_coca_2}
\end{equation}

Clearly, if we impose the constraint $q_{2j}=q_{2j+1}, \forall j\in [0,d/2-1]$, the vector in Equation (\ref{theorem_dual_coca_2}) becomes null and we deduce that:
\begin{equation}
\small
\begin{aligned}
&f(\mathbf{q}_m,n)\circ \mathbf{t}_n-\mathbf{q}_m\circ f(\mathbf{t}_n,n) = \mathbf{0}
\label{eq:17}
\end{aligned}
\end{equation}

Consequently, with the constraint $q_{2j}=q_{2j+1}, \forall j\in [0,d/2-1]$, we have:
\begin{equation}
\small
\begin{aligned}
& \text{\rm{Re}}(\langle f(\mathbf{q}_m,m), \mathbf{q}_m \circ f(\mathbf{t}_n,n)  \rangle) \\
&= \text{Re}(\langle f(\mathbf{q}_m,m), f(\mathbf{q}_m,n) \circ \mathbf{t}_n \rangle)
\label{theorem_dual_coca_3}
\end{aligned}
\end{equation}

\noindent\textit{Step 2.} We further demonstrate that, $q_{2j}=q_{2j+1}, \forall j\in [0,d/2-1]$ is in fact a redundant constraint when calculating $\text{Re}(\langle f(\mathbf{q}_m,m), f(\mathbf{q}_m,n) \circ \mathbf{t}_n \rangle)$. To verify this, we expand the inner product:
\begin{equation}
\small
\begin{aligned}
&\text{Re}(\langle f(\mathbf{q}_m,m), f(\mathbf{q}_m,n) \circ \mathbf{t}_n \rangle)\\
       &=\sum_{j=0}^{d/2-1}[(q_{2j}^2t_{2j}+q_{2j+1}^2t_{2j+1})\cos((m-n)\theta_j) \\
       &\quad +(q_{2j}q_{2j+1}t_{2j}-q_{2j+1}q_{2j}t_{2j+1})\sin((m-n)\theta_j)] \\
\end{aligned}
\end{equation}
Recall again $t_{2j}=t_{2j+1}, \forall j\in [0,d/2-1]$, we have
\begin{equation}
\small
\begin{aligned}
& \text{Re}(\langle f(\mathbf{q}_m,m), f(\mathbf{q}_m,n) \circ \mathbf{t}_n \rangle)\\
&=\sum_{j=0}^{d/2-1} t_{2j}[(q_{2j}^2+q_{2j+1}^2)\cos((m-n)\theta_j)] \\
       &=\sum_{j=0}^{d/2-1} t_{2j} |\mathbf{q}_j|^2\cos((m-n)\theta_j)
\end{aligned}
\label{eq:magnitude dependence}
\end{equation}

This implies that $\text{Re}(\langle f(\mathbf{q}_m,m), f(\mathbf{q}_m,n) \circ \mathbf{t}_n \rangle)$ depends solely on the magnitude of $\mathbf{q}_j=(q_{2j},q_{2j+1})$ in 2-D subspace, demonstrating the independence of the relationship between $q_{2j}$ and $q_{2j+1}$. Refer to Appendix \ref{Homeomorphism of representation space} for the rigorous proof.

Now we conclude that, with the constraint $q_{2j}=q_{2j+1}, \forall j\in [0,d/2-1]$, $\text{\rm{Re}}(\langle f(\mathbf{q}_m,m), \mathbf{q}_m \circ f(\mathbf{t}_n,n)  \rangle)$  is equivalent to $\text{Re}(\langle f(\mathbf{q}_m,m), f(\mathbf{q}_m,n) \circ \mathbf{t}_n \rangle)$ with no constraint on query. \qedsymbol

By removing $q_{2j} = q_{2j+1}$ constraint, we designate this modified version as CoCA-Slack. The mathematical definition is provided in Appendix \ref{Slack Position Embedding}.

\section{Experimental Setting}
\label{experimental setting}
This section provides an overview of the experimental setup, including details regarding the training data utilized and the baseline models employed to evaluate the effectiveness of the proposed method.

\subsection{Training Data}
\label{Training Data}
Our model undergoes training on a combination of datasets, including the Pile training dataset~\citep{Gao2020ThePA}, BookCorpus ~\citep{Zhu2015AligningBA}, and the Wikipedia Corpus~\citep{wiki}. Additionally, we integrate manually collected open-source code from GitHub repositories with at least 1 star. From these datasets, we derive a sample of approximately 50B tokens, maintaining a composition of 75\% text and 25\% code.

\subsection{Model Variants}
\label{Model Variants}
To evaluate the effectiveness of our proposed approach, we  train 3  models from scratch under identical experimental settings, including  ALibi \citep{Press2021TrainST}, RoFomer \citep{Su2021RoFormerET}, and RoFormer+CoCA.  All models share common specifications, featuring a size of 350M, 24 layers, a hidden dimension of 1024, 16 attention heads, and a maximum sequence length of 512. The key distinctions among them lie in variations in self-attention mechanisms and position embeddings. The implementation is optimized based on EleutherAI GPT-NeoX\footnote{\href{https://github.com/EleutherAI/gpt-neox/tree/v2.0}{https://github.com/EleutherAI/gpt-neox/tree/v2.0}}. Training a model from scratch demands substantial computational resources. Therefore, we also conduct experiments involving fine-tuning existing LLMs with a drop-in CoCA module. For this purpose, we utilize the LLaMA-7B model \citep{Touvron2023LLaMAOA}, which was trained with a context length of 2,048. Additionally, we employ dynamic-NTK for all the above models. 

In summary, our comparison models are categorized as follows: ALibi, RoFormer, RoFormer+CoCA, RoFormer+dynamic NTK, and RoFormer+dynamic NTK \& CoCA, all falling under the \textit{training from scratch} category. Meanwhile, LLaMA-7B, LLaMA-7B+CoCA, LLaMA-7B+dynamic NTK, and LLaMA-7B+dynamic NTK \& CoCA belong to the \textit{fine-tuning LLM with drop-in CoCA} category.

\subsection{Implementation Detials}
\noindent \textbf{Pre-training Procedure}  We train all models using the next token prediction objective. We use AdamW ~\citep{Loshchilov2017FixingWD} with $\beta_1$ = 0.9 and $\beta_2$ = 0.95. The learning rate follows a linear warm-up of 1\% of total steps, starting from 1e-7. Subsequently, the learning rate is adjusted to 1e-4 with linear decay, eventually reaching 1e-5. The training utilizes 8 A100 GPUs, with a global batch size of 256 and 2 gradient steps accumulation, taking approximately 96 hours for 2 epochs.

\noindent \textbf{Fine-tuning Procedure}
To integrate CoCA in LLaMA, we employ a three-stage fine-tuning strategy: 
(1) only updating the $K$ projection (7\% of parameters). This stage aims to reconstruct the $K$ projection in CoCA. By freezing the other parameters, we maintain attention scores as closely as possible to those of vanilla self-attention. 
(2) updating the $QKV$ projection (21\% of parameters). This stage aims to address intrinsic over-fitting in vanilla self-attention caused by undesired behaviors between RoPE and attention matrices. 
(3) fine-tuning all parameters. Each stage involves 15K steps, totaling 7.5B tokens (22B tokens overall), using the next token prediction objective. The training length of LLaMA-7B + CoCA remains at 2,048 as in the original model. 
All experiments are conducted with 32 A100 GPUs, setting a per-device batch size to 8 without gradient accumulation. 

\section{Experiment Results}
\label{Experiment Results}

We conducted experiments to shed light on the following reasonable doubts:
\begin{itemize}[itemsep= 0.2pt,topsep = 0.2pt,partopsep=0.2pt]
    \item Can our new attention mechanism CoCA improve the long context extrapolation performance of existing models?
    \item Can combining CoCA with other extending methods for RoPE effectively solve the three types of rotational boundary problems discussed in Appendix \ref{Rotary Borders Analysis}?
\end{itemize}
\subsection{Long Sequence Language Modeling}

\begin{table*}[ht!]
\setstretch{1.25}
\centering
\small
\begin{tabular*}{\textwidth}{@{\extracolsep{\fill}}lcccccccc}
\toprule
\multirow{2}{*}{\textbf{Method}} & \multicolumn{7}{c}{\textbf{Evaluation Context   Window Size (Perplexity  ↓)}}                                         \\ \cline{2-8} 
                                 & \textbf{512}   & \textbf{1024}  & \textbf{2048}  & \textbf{4096}  & \textbf{8192}  & \textbf{16k}   & \textbf{32k}    \\ \hline
\multicolumn{8}{c}{\textit{Training model from scratch}}                                                                                                              \\ \hline
ALibi                            & \textbf{18.69} & 21.27          & 28.20          & {35.66} & \textbf{37.03} & OOM              & OOM               \\
RoFomer                          & 19.66          & 411.50         & 3276.00        & 3026.00        & 3028.00        & inf            & inf             \\
+ dynamic NTK                    & 19.66          & 22.30          & 38.00          & 75.75          & 138.13         & 370.75         & 380.75          \\
+ CoCA                           & 20.11          & 33.47          & 69.06          & 113.19         & 157.38         & 141.00         & 171.63          \\
+ dynamic NTK \& CoCA            & 20.11          & \textbf{20.81} & \textbf{25.88} & \textbf{34.16}          & 55.75          & \textbf{89.31} & \textbf{101.13} \\ 
\hline
\multicolumn{8}{c}{\textit{Fine-tuning LLM with drop-in CoCA}}                                                                                                          \\ \hline
LLaMA-7B                         & \textbf{9.25}  & \textbf{7.56}  & \textbf{7.30}  & 9673.14        & inf            & inf            & inf             \\
+ dynamic NTK                    & {9.25}  & {7.56}  & {7.30}  & 9.40           & 14.40          & 63.62          & 133.87          \\
+ CoCA                           & 9.91           & 8.49           & 8.27           & 24.23          & 42.00          & 23.83          & 29.95           \\
+ dynamic NTK \& CoCA            & 9.91           & 8.49           & 8.27           & \textbf{8.61}  & \textbf{9.56}  & \textbf{11.10} & \textbf{13.98}  \\
\bottomrule
\end{tabular*}
\caption{Evaluation perplexity on 100 PG-19 documents using sliding window (S = 512) strategy. Dynamic-NTK is employed without fine-tuning. The best result is highlighted in bold.}
\label{tab:1}
\end{table*}

We evaluate the long sequence language modeling performance of both our model and baseline models on the test splits of the PG-19 dataset \citep{Rae2019CompressiveTF}. For this evaluation, we randomly select a subsample comprising 100 documents, each containing at least 32,768 SentencePiece \citep{kudo2018sentencepiece} tokens. We then truncate each test document to its initial 32,768 tokens. The evaluation involves calculating perplexity across different context window sizes using a sliding window approach, as described by \citep{Press2021TrainST}, with a stride of 512. The perplexity results for both our models and baselines are presented in Table \ref{tab:1} and Figure \ref{fig:ppl}. 

Based on our experiments, the evaluation results indicate that models combined with CoCA exhibit significantly improved perplexity with longer inference sequence length. For pre-trained models, by increasing the context window size from 512 (training context window size) to 32k, the perplexity of CoCA only increases from 20.11 to 171.63, whereas the perplexity of RoFormer becomes $\inf$. Additionally, by increasing the context window size from 2K to 32K, the perplexity of fine-tuned LLaMA-7B+CoCA only increases 21.68, while LLaMA-7B with other extending methods increases more than 100. In general, we observe a consistent trend of CoCA achieving better perplexity with longer context windows. This suggests that CoCA has a more robust position embedding, enabling it to handle long context more effectively.


In contrast, we observe that models extended through the direct application of dynamic NTK-aware Scaled RoPE exhibit a larger increase in perplexity at longer sequences. The perplexity of both RoFormer+dynamic NTK and LLaMA-7B+dynamic NTK remains significantly higher than that combining CoCA. This difference becomes more pronounced as the sequence length increases. When the inference sequence length reaches 32k, the perplexity of RoFormer+dynamic NTK increases to 380.75, while the result for RoFormer+CoCA is only 171.63. Similarly, the perplexity of LLaMA-7B+dynamic NTK reaches 133.87, whereas LLaMA-7B+CoCA is only 29.95.

It is worth noting that the model achieves the best performance when both dynamic NTK and CoCA are combined. Particularly, LLaMA-7B+dynamic NTK \& CoCA consistently maintains a very low perplexity. Even when the inference sequence length has reached 32k (16 $\times$ longer than the training length), the perplexity is only 13.89. This indicates that combining CoCA with other extending methods for RoPE can effectively address the three types of rotational boundary problems, achieving robust long-text extrapolation modeling capabilities.

\subsection{Long Context Retrieval}
\label{Long Context Retrieval}

\begin{table*}[ht!]
\setstretch{1.25}
\centering
\small
\begin{tabular*}{\textwidth}{@{\extracolsep{\fill}}lcccccccc}
\toprule
\multirow{2}{*}{\textbf{Method}} & \multicolumn{7}{c}{\textbf{Evaluation Context Window Size (Accuracy↑)}}                                         \\ \cline{2-8} 
                                 & \textbf{512}  & \textbf{1024} & \textbf{2048} & \textbf{4096} & \textbf{8192} & \textbf{16k}   & \textbf{32k}   \\ \hline
\multicolumn{8}{c}{\textit{Traning model from scratch}}                                                                                                        \\ \hline
ALibi                            & {0.82} & 0.65          & 0.28          & {0.18} & {0.12} & OOM              & OOM              \\
RoFomer                          & 0.99          & 0.53          & 0.30          & 0.18          & 0.04          & 0.02           & 0.04           \\
+ dynamic NTK                    & 0.99          & 1.00          & 0.95          & 0.70          & 0.41          & 0.16           & 0.06           \\
+ CoCA                           & 1.00          & 0.64          & 0.33          & 0.19          & 0.06          & 0.02           & 0.04           \\
+ dynamic NTK \& CoCA            & \textbf{1.00} & \textbf{1.00} & \textbf{0.96} & \textbf{0.89} & \textbf{0.50} & \textbf{0.23}  & \textbf{0.08}  \\ \hline
\multicolumn{8}{c}{\textit{Fine-tuning LLM with drop-in CoCA}}                                                                                                    \\ \hline
LLaMA-7B                         & {1.00} & {1.00} & {1.00} & 0.61          & 0.21          & 0.07           & 0.09           \\
+ dynamic NTK                    & {1.00} & {1.00} & {1.00} & 0.81          & 0.26          & 0.06           & 0.03           \\
+ CoCA                           & 1.00          & 1.00          & 1.00          & 0.71          & 0.28          & 0.11           & 0.10           \\
+ dynamic NTK \& CoCA            & \textbf{1.00}          & \textbf{1.00}          & \textbf{1.00}         & \textbf{1.00} & \textbf{0.85} & \textbf{0.51} & \textbf{0.30} \\
\bottomrule
\end{tabular*}
\caption{Long context retrieval performance on passkey retrieval task. The best result is highlighted in bold.}
\label{tab:2}
\end{table*}

Perplexity evaluates the performance of language model in predicting the next token. However, it is insufficient for a comprehensive assessment of the effective context window size. To address this, we conducted experiments using a passkey retrieval task \citep{DBLP:journals/corr/abs-2305-16300} to evaluate our method and baselines. The task involves identifying and retrieving a randomly hidden passkey within a lengthy document. More details of task definition and test sample generation settings can be found in Appendix \ref{passkey task}. Table \ref{tab:2} illustrates the accuracy of all tested models and their variants. 

It is evident that ALibi exhibited failures when tested on sequences that were 1$\times$ longer than its training length, attributed to its local hypothesis. In contrast, our model consistently demonstrated superior accuracy. RoFormer+dynamic NTK \& CoCA maintained a 50\% accuracy, even with the test sequence length expanded to 16$\times$ its training length. Similarly, LLaMA-7B+dynamic NTK \& CoCA still maintained a 30\% accuracy when the test length was up to 32K.

\subsection{Impact of Strict and Slack Constraint on Q}
\label{sec:strict-and-slack-constraint-of-q}

\begin{table*}[ht!]
\setstretch{1.25}
  \centering
  \small
  \begin{tabular*}{\textwidth}{@{\extracolsep{\fill}}ccccccccc}
    \toprule
    \multicolumn{2}{c}{\textbf{Method}} &
      \multicolumn{1}{l}{\textbf{512}} &
      \multicolumn{1}{l}{\textbf{1024}} &
      \multicolumn{1}{l}{\textbf{2048}} &
      \multicolumn{1}{l}{\textbf{4096}} &
      \multicolumn{1}{l}{\textbf{8192}} &
      \multicolumn{1}{l}{\textbf{16384}} &
      \multicolumn{1}{l}{\textbf{32768}} \\ \hline
    \multicolumn{9}{c}{\textit{Performance on Long Sequence Modeling (Perplexity)}} \\ \hline
    &
      CoCA-Slack &
      {\color[HTML]{000000} 20.11} &
      19.02 &
      24.92 &
      40.53 &
      68.38 &
      92.75 &
      103.44 \\
    \multirow{-2}{*}{ntk-2} &
      CoCA-Strict &
      {\color[HTML]{FF0000} +0.07} &
      {\color[HTML]{FF0000} +0.61} &
      {\color[HTML]{00B050} -1.58} &
      {\color[HTML]{00B050} -4.03} &
      {\color[HTML]{FF0000} +15.37} &
      {\color[HTML]{FF0000} +12.38} &
      {\color[HTML]{FF0000} +1.94} \\
    &
      CoCA-Slack &
      {\color[HTML]{000000} 20.11} &
      20.81 &
      25.88 &
      34.16 &
      55.75 &
      89.31 &
      101.13 \\
    \multirow{-2}{*}{ntk-4} &
      CoCA-Strict &
      {\color[HTML]{FF0000} +0.07} &
      {\color[HTML]{00B050} -0.49} &
      {\color[HTML]{00B050} -0.66} &
      {\color[HTML]{00B050} -0.88} &
      {\color[HTML]{FF0000} +3.16} &
      {\color[HTML]{00B050} -18.25} &
      {\color[HTML]{00B050} -2.57} \\
    &
      CoCA-Slack &
      {\color[HTML]{000000} 20.11} &
      23.66 &
      29.05 &
      37.47 &
      55.5 &
      88.88 &
      111.38 \\
    \multirow{-2}{*}{ntk-8} &
      CoCA-Strict &
      {\color[HTML]{FF0000} +0.07} &
      {\color[HTML]{00B050} -1.74} &
      {\color[HTML]{00B050} -0.64} &
      {\color[HTML]{FF0000} +1.16} &
      {\color[HTML]{FF0000} +0.03} &
      {\color[HTML]{FF0000} +0.5} &
      {\color[HTML]{FF0000} +0.31} \\ \hline
    \multicolumn{9}{c}{\textit{Performance on Long Context Retrieval (Passkey Accuracy)}} \\ \hline
    &
      CoCA-Slack &
      {\color[HTML]{000000} 1.0} &
      0.99 &
      0.94 &
      0.77 &
      0.47 &
      0.27 &
      0.15 \\
    \multirow{-2}{*}{ntk-2} &
      CoCA-Strict &
      {\color[HTML]{00B050} +0.0} &
      {\color[HTML]{FF0000} -0.12} &
      {\color[HTML]{FF0000} -0.3} &
      {\color[HTML]{FF0000} -0.42} &
      {\color[HTML]{FF0000} -0.34} &
      {\color[HTML]{FF0000} -0.22} &
      {\color[HTML]{FF0000} -0.07} \\
    &
      
      CoCA-Slack &
      {\color[HTML]{000000} 1.0} &
      1.0 &
      0.96 &
      0.89 &
      0.5 &
      0.23 &
      0.08 \\
    \multirow{-2}{*}{ntk-4} &
      CoCA-Strict &
      {\color[HTML]{00B050} +0.0} &
      {\color[HTML]{FF0000} -0.11} &
      {\color[HTML]{FF0000} -0.38} &
      {\color[HTML]{FF0000} -0.46} &
      {\color[HTML]{FF0000} -0.38} &
      {\color[HTML]{FF0000} -0.19} &
      {\color[HTML]{FF0000} -0.02} \\
    &
      CoCA-Slack &
      {\color[HTML]{000000} 1.0} &
      0.98 &
      0.99 &
      0.85 &
      0.5 &
      0.11 &
      0.02 \\
    \multirow{-2}{*}{ntk-8} &
      CoCA-Strict &
      {\color[HTML]{00B050} +0.0} &
      {\color[HTML]{FF0000} -0.05} &
      {\color[HTML]{FF0000} -0.34} &
      {\color[HTML]{FF0000} -0.51} &
      {\color[HTML]{FF0000} -0.4} &
      {\color[HTML]{FF0000} -0.07} &
      {\color[HTML]{FF0000} -0.01} \\ \bottomrule
  \end{tabular*}
  \caption{Comparison results for the Strict and Slack Constraints of Q in our proposed CoCA module. Superior performance to CoCA-Slack is indicated by the green color, while inferior performance is signified by the red color. The perplexity of the strict and slack models is comparable, whereas the strict model achieved lower accuracy in the passkey retrieval task.}
  \label{tb:12}
\end{table*}

As mentioned in Section \ref{Slacking the constraint on query}, we implement a slack version of CoCA, referred to as CoCA-Slack.  In this section, under the same experimental settings, we implement two versions of CoCA based on RoFormer-350M, labeled as CoCA-Slack and CoCA-Strict. The comparison results between them are shown in Table \ref{tb:12}.

We observe that the CoCA-Strict and CoCA-Slack models exhibit similar performance in long sequence language modeling, as evidenced by comparable perplexity results. However, in the passkey retrieval task, contrary to our initial expectations, the CoCA-Strict model produces significantly lower results. This unexpected outcome suggests that models with a slack constraint may offer additional performance advantages, such as a larger effective context window size. 

Understanding the reasons behind the superiority of slack constraints will be a key focus of our future work. In this regard, we provide some theoretical insights in Appendices \ref{Homeomorphism of representation space} and \ref{Slack Position Embedding}. These insights aim to shed light on the underlying mechanisms that contribute to the observed differences and lay the groundwork for a more comprehensive analysis in subsequent research.

\section{Conclusion}
In this paper, we introduce Collinear Constrained Attention (CoCA), a novel approach that integrates position embedding with the self-attention mechanism. This innovation addresses undesired behaviors occurring around the context window boundary, which stem from discrepancies between RoPE and attention matrices. To the best of our knowledge, we are the first to analyze the initial angles between queries and keys in the self-attention mechanism, which gives rise to anomalous phenomena in RoPE. Furthermore, we deduce a slack constraint for our implementation of CoCA. Extensive experiments demonstrate that incorporating CoCA into existing models significantly enhances performance in both long sequence language modeling and long context retrieval tasks. Additionally, the simultaneous integration of CoCA with other extended RoPE methods (e.g., dynamic-NTK) effectively mitigates three types of rotation boundary issues, resulting in remarkably improved capabilities for long context extrapolation.

\section*{Limitations}
Our current approach, CoCA, has thus far undergone exclusive validation on RoPE. Experimental results demonstrate that CoCA enhances the long-context extrapolation performance of LLMs and further augments other extension methods by addressing rotational boundary issues. However, questions arise regarding its applicability to more general methods. While the effectiveness of slack position embedding (SPE) is evident, a deeper understanding of the underlying reasons for its superior performance necessitates further investigation.

\bibliography{acl_latex}

\begin{thebibliography}{29}
\expandafter\ifx\csname natexlab\endcsname\relax\def\natexlab#1{#1}\fi

\bibitem[{a.~Smith and Gray(2018)}]{Smith2018}
Daniel~G. a.~Smith and Johnnie Gray. 2018.
\newblock \href {https://doi.org/10.21105/joss.00753} {opt\_einsum - a python package for optimizing contraction order for einsum-like expressions}.
\newblock \emph{Journal of Open Source Software}, 3(26):753.

\bibitem[{Bai et~al.(2023)Bai, Bai, Chu et~al.}]{qwen}
Jinze Bai, Shuai Bai, Yunfei Chu, et~al. 2023.
\newblock Qwen technical report.
\newblock \emph{arXiv preprint arXiv:2309.16609}.

\bibitem[{Beltagy et~al.(2020)Beltagy, Peters, and Cohan}]{beltagy2020longformer}
Iz~Beltagy, Matthew~E Peters, and Arman Cohan. 2020.
\newblock \href {https://arxiv.org/abs/2004.05150} {Longformer: The long-document transformer}.
\newblock \emph{arXiv preprint arXiv:2004.05150}.

\bibitem[{Black et~al.(2022)Black, Biderman, Hallahan et~al.}]{black-etal-2022-gpt}
Sidney Black, Stella Biderman, Eric Hallahan, et~al. 2022.
\newblock \href {https://doi.org/10.18653/v1/2022.bigscience-1.9} {{GPT}-{N}eo{X}-20{B}: An open-source autoregressive language model}.
\newblock In \emph{Proceedings of BigScience Episode {\#}5 -- Workshop on Challenges {\&} Perspectives in Creating Large Language Models}, pages 95--136, virtual+Dublin. Association for Computational Linguistics.

\bibitem[{bloc97(2023)}]{NTK}
bloc97. 2023.
\newblock \href {https://www.reddit.com/r/LocalLLaMA/comments/14lz7j5/ntkaware\_scaled\_rope\_allows\_llama\_modes\_to\_have/.} {Ntk-aware scaled rope allows llama models to have extended (8k+) context size without any fine-tuning and minimal perplexity degradation}.

\bibitem[{Chen et~al.(2023)Chen, Wong, Chen, and Tian}]{Chen2023ExtendingCW}
Shouyuan Chen, Sherman Wong, Liangjian Chen, and Yuandong Tian. 2023.
\newblock \href {https://api.semanticscholar.org/CorpusID:259262376} {Extending context window of large language models via positional interpolation}.
\newblock \emph{ArXiv}, abs/2306.15595.

\bibitem[{Chi et~al.(2022)Chi, Fan, Ramadge, and Rudnicky}]{Chi2022KERPLEKR}
Ta{-}Chung Chi, Ting{-}Han Fan, Peter~J. Ramadge, and Alexander Rudnicky. 2022.
\newblock \href {http://papers.nips.cc/paper\_files/paper/2022/hash/37a413841a614b5414b333585e7613b8-Abstract-Conference.html} {{KERPLE:} kernelized relative positional embedding for length extrapolation}.
\newblock In \emph{Advances in Neural Information Processing Systems 35: Annual Conference on Neural Information Processing Systems 2022, NeurIPS 2022, New Orleans, LA, USA, November 28 - December 9, 2022}.

\bibitem[{Contributors(2023)}]{2023opencompass}
OpenCompass Contributors. 2023.
\newblock Opencompass: A universal evaluation platform for foundation models.
\newblock \url{https://github.com/open-compass/opencompass}.

\bibitem[{Dao et~al.(2022)Dao, Fu, Ermon et~al.}]{dao2022flashattention}
Tri Dao, Daniel~Y. Fu, Stefano Ermon, et~al. 2022.
\newblock Flash{A}ttention: Fast and memory-efficient exact attention with {IO}-awareness.
\newblock In \emph{Advances in Neural Information Processing Systems}.

\bibitem[{Ding et~al.(2023)Ding, Ma, Dong et~al.}]{ding2023longnet}
Jiayu Ding, Shuming Ma, Li~Dong, et~al. 2023.
\newblock \href {https://arxiv.org/abs/2307.02486} {Longnet: Scaling transformers to 1,000,000,000 tokens}.
\newblock \emph{arXiv preprint arXiv:2307.02486}.

\bibitem[{Emozilla(2023)}]{dynamicNTK}
Emozilla. 2023.
\newblock \href {https://www.reddit.com/r/LocalLLaMA/comments/14mrgpr/dynamically_scaled_rope_further_increases/.} {Dynamically scaled rope further increases performance of long context llama with zero fine-tuning}.

\bibitem[{Foundation(2021)}]{wiki}
Wikimedia Foundation. 2021.
\newblock \href {https://dumps.wikimedia.org} {Wikimedia downloads}.

\bibitem[{Gao et~al.(2020)Gao, Biderman, Black et~al.}]{Gao2020ThePA}
Leo Gao, Stella~Rose Biderman, Sid Black, et~al. 2020.
\newblock \href {https://api.semanticscholar.org/CorpusID:230435736} {The pile: An 800gb dataset of diverse text for language modeling}.
\newblock \emph{ArXiv}, abs/2101.00027.

\bibitem[{Han et~al.(2023)Han, Wang, Xiong et~al.}]{han2023lm}
Chi Han, Qifan Wang, Wenhan Xiong, et~al. 2023.
\newblock \href {https://arxiv.org/pdf/2308.16137} {Lm-infinite: Simple on-the-fly length generalization for large language models}.
\newblock \emph{arXiv preprint arXiv:2308.16137}.

\bibitem[{Huang et~al.(2023)Huang, Xu, Jiang et~al.}]{huang2023advancing}
Yunpeng Huang, Jingwei Xu, Zixu Jiang, et~al. 2023.
\newblock \href {https://arxiv.org/abs/2311.12351} {Advancing transformer architecture in long-context large language models: A comprehensive survey}.
\newblock \emph{arXiv preprint arXiv:2311.12351}.

\bibitem[{Kudo and Richardson(2018)}]{kudo2018sentencepiece}
Taku Kudo and John Richardson. 2018.
\newblock \href {https://doi.org/10.18653/V1/D18-2012} {Sentencepiece: {A} simple and language independent subword tokenizer and detokenizer for neural text processing}.
\newblock In \emph{Proceedings of the 2018 Conference on Empirical Methods in Natural Language Processing, {EMNLP} 2018: System Demonstrations, Brussels, Belgium, October 31 - November 4, 2018}, pages 66--71. Association for Computational Linguistics.

\bibitem[{Loshchilov and Hutter(2017)}]{Loshchilov2017FixingWD}
Ilya Loshchilov and Frank Hutter. 2017.
\newblock \href {https://api.semanticscholar.org/CorpusID:3312944} {Fixing weight decay regularization in adam}.
\newblock \emph{ArXiv}, abs/1711.05101.

\bibitem[{Mohtashami and Jaggi(2023)}]{DBLP:journals/corr/abs-2305-16300}
Amirkeivan Mohtashami and Martin Jaggi. 2023.
\newblock \href {https://doi.org/10.48550/ARXIV.2305.16300} {Landmark attention: Random-access infinite context length for transformers}.
\newblock \emph{CoRR}, abs/2305.16300.

\bibitem[{Peng et~al.(2023)Peng, Quesnelle, Fan, and Shippole}]{DBLP:journals/corr/abs-2309-00071}
Bowen Peng, Jeffrey Quesnelle, Honglu Fan, and Enrico Shippole. 2023.
\newblock \href {https://doi.org/10.48550/ARXIV.2309.00071} {Yarn: Efficient context window extension of large language models}.
\newblock \emph{CoRR}, abs/2309.00071.

\bibitem[{Press et~al.(2022)Press, Smith, and Lewis}]{Press2021TrainST}
Ofir Press, Noah~A. Smith, and Mike Lewis. 2022.
\newblock \href {https://openreview.net/forum?id=R8sQPpGCv0} {Train short, test long: Attention with linear biases enables input length extrapolation}.
\newblock In \emph{The Tenth International Conference on Learning Representations, {ICLR} 2022, Virtual Event, April 25-29, 2022}. OpenReview.net.

\bibitem[{Rae et~al.(2020)Rae, Potapenko, Jayakumar, Hillier, and Lillicrap}]{Rae2019CompressiveTF}
Jack~W. Rae, Anna Potapenko, Siddhant~M. Jayakumar, Chloe Hillier, and Timothy~P. Lillicrap. 2020.
\newblock \href {https://openreview.net/forum?id=SylKikSYDH} {Compressive transformers for long-range sequence modelling}.
\newblock In \emph{8th International Conference on Learning Representations, {ICLR} 2020, Addis Ababa, Ethiopia, April 26-30, 2020}. OpenReview.net.

\bibitem[{Ruder et~al.(2019)Ruder, Peters, Swayamdipta, and Wolf}]{DBLP:conf/naacl/RuderPSW19}
Sebastian Ruder, Matthew~E. Peters, Swabha Swayamdipta, and Thomas Wolf. 2019.
\newblock \href {https://doi.org/10.18653/V1/N19-5004} {Transfer learning in natural language processing}.
\newblock In \emph{Proceedings of the 2019 Conference of the North American Chapter of the Association for Computational Linguistics: Human Language Technologies, {NAACL-HLT} 2019, Minneapolis, MN, USA, June 2, 2019, Tutorial Abstracts}, pages 15--18. Association for Computational Linguistics.

\bibitem[{Su et~al.(2024)Su, Ahmed, Lu, Pan, Bo, and Liu}]{Su2021RoFormerET}
Jianlin Su, Murtadha H.~M. Ahmed, Yu~Lu, Shengfeng Pan, Wen Bo, and Yunfeng Liu. 2024.
\newblock \href {https://doi.org/10.1016/J.NEUCOM.2023.127063} {Roformer: Enhanced transformer with rotary position embedding}.
\newblock \emph{Neurocomputing}, 568:127063.

\bibitem[{Sun et~al.(2023)Sun, Dong, Patra, Ma, Huang, Benhaim, Chaudhary, Song, and Wei}]{Sun2022ALT}
Yutao Sun, Li~Dong, Barun Patra, Shuming Ma, Shaohan Huang, Alon Benhaim, Vishrav Chaudhary, Xia Song, and Furu Wei. 2023.
\newblock \href {https://doi.org/10.18653/V1/2023.ACL-LONG.816} {A length-extrapolatable transformer}.
\newblock In \emph{Proceedings of the 61st Annual Meeting of the Association for Computational Linguistics (Volume 1: Long Papers), {ACL} 2023, Toronto, Canada, July 9-14, 2023}, pages 14590--14604. Association for Computational Linguistics.

\bibitem[{Touvron et~al.(2023{\natexlab{a}})Touvron, Lavril, Izacard et~al.}]{Touvron2023LLaMAOA}
Hugo Touvron, Thibaut Lavril, Gautier Izacard, et~al. 2023{\natexlab{a}}.
\newblock \href {https://api.semanticscholar.org/CorpusID:257219404} {Llama: Open and efficient foundation language models}.
\newblock \emph{ArXiv}, abs/2302.13971.

\bibitem[{Touvron et~al.(2023{\natexlab{b}})Touvron, Martin, Stone et~al.}]{Touvron2023Llama2O}
Hugo Touvron, Louis Martin, Kevin~R. Stone, et~al. 2023{\natexlab{b}}.
\newblock \href {https://api.semanticscholar.org/CorpusID:259950998} {Llama 2: Open foundation and fine-tuned chat models}.
\newblock \emph{ArXiv}, abs/2307.09288.

\bibitem[{Vaswani et~al.(2017)Vaswani, Shazeer, Parmar, Uszkoreit, Jones, Gomez, Kaiser, and Polosukhin}]{vaswani2017attention}
Ashish Vaswani, Noam Shazeer, Niki Parmar, Jakob Uszkoreit, Llion Jones, Aidan~N. Gomez, Lukasz Kaiser, and Illia Polosukhin. 2017.
\newblock \href {https://proceedings.neurips.cc/paper/2017/hash/3f5ee243547dee91fbd053c1c4a845aa-Abstract.html} {Attention is all you need}.
\newblock In \emph{Advances in Neural Information Processing Systems 30: Annual Conference on Neural Information Processing Systems 2017, December 4-9, 2017, Long Beach, CA, {USA}}, pages 5998--6008.

\bibitem[{Xiao et~al.(2023)Xiao, Tian, Chen et~al.}]{xiao2023efficient}
Guangxuan Xiao, Yuandong Tian, Beidi Chen, et~al. 2023.
\newblock \href {https://arxiv.org/pdf/2309.17453} {Efficient streaming language models with attention sinks}.
\newblock \emph{arXiv preprint arXiv:2309.17453}.

\bibitem[{Zhu et~al.(2015)Zhu, Kiros, Zemel, Salakhutdinov, Urtasun, Torralba, and Fidler}]{Zhu2015AligningBA}
Yukun Zhu, Ryan Kiros, Richard~S. Zemel, Ruslan Salakhutdinov, Raquel Urtasun, Antonio Torralba, and Sanja Fidler. 2015.
\newblock \href {https://doi.org/10.1109/ICCV.2015.11} {Aligning books and movies: Towards story-like visual explanations by watching movies and reading books}.
\newblock In \emph{2015 {IEEE} International Conference on Computer Vision, {ICCV} 2015, Santiago, Chile, December 7-13, 2015}, pages 19--27. {IEEE} Computer Society.

\end{thebibliography}

\appendix

\section{Related Work}
Existing researches are mainly focused on the submodule of attention kernel or position embedding \citep{huang2023advancing}. In the following sections, we will separately introduce works on these two aspects: Section \ref{Efficient Attention Mechanisms} primarily addresses the former, while Section \ref{Extrapolative Positional Embedding Methods} delves into the latter.

\subsection{Efficient Attention Mechanisms}
\label{Efficient Attention Mechanisms}

Several works aim to implement efficient attention mechanisms with reduced computational demands, even achieving linear complexity. This enables extending the effective context length boundary of LLMs during inference by directly increasing \(L_{max}\) in the pre-training stage \citep{ding2023longnet, DBLP:journals/corr/abs-2305-16300}.  Noteworthy approaches include Longformer \citep{beltagy2020longformer}, utilizing slide window attention, and models such as StreamingLLM \citep{xiao2023efficient} and LM-Infinite \citep{han2023lm}, which leverage a global-local attention mechanism. 
These variants have achieved success to a certain extent, but still face issues we unveiled in this work when using RoPE as their positional encoding method.


\subsection{Extrapolative Position Embedding Methods}
\label{Extrapolative Positional Embedding Methods}

Extrapolative position embedding methods aim to enhance the length generalization capability of LLMs. 

\subsubsection{Attention Bias}
In seeking alternatives to the explicit encoding of positional information, researchers have explored the integration of attention bias to capture the sequential and temporal nuances inherent in natural language. 
Early approaches, such as T5 \citep{DBLP:conf/naacl/RuderPSW19}, incorporate learnable attention bias. 
However, these methods do not explicitly address the challenge of length extrapolation. 
ALibi \citep{Press2021TrainST} introduces a negative causal attention bias in a heuristic manner. 
Extending the ALiBi-style attention bias, KERPLE \citep{Chi2022KERPLEKR} treats it as a composition triangle kernel for self-attention and modifies style Xpos \citep{Sun2022ALT} by integrating it with RoPE. 
While these approaches effectively manage to maintain low perplexity levels, they fall short in capturing long-range dependencies due to introducing local hypotheses to context tokens.

\subsubsection{Extend RoPE}
Besides, various strategies have been explored to extend RoPE \citep{Su2021RoFormerET}, a commonly employed positional encoding method in popular LLMs. 
Recent approaches involve simply scaling it to extrapolate the inference context length with minimal or no fine-tuning.  
For instance, Position Interpolation (PI) ~\citep{Chen2023ExtendingCW} applies linear scaling on each position number from \(n\) to \(n/k\), densifying the representation space to extend the farthest length boundary by \(k\) times. 
Other approaches, such as NTK-aware Scaled RoPE \citep{NTK} and Dynamic-NTK \citep{dynamicNTK}, combine high-frequency extrapolation and low-frequency interpolation. 
These training-free methods require limited code changes during inference \citep{DBLP:journals/corr/abs-2309-00071}. 
However, these methods aim solely at alleviating the problem of modeling the rotation angles in out-of-distribution (OOD) positions without recognizing the intrinsic correlation between attention matrices and rotation angles. Therefore, these methods still suffer from a limited context window extending ratio.

Previous methods independently investigate self-attention and position embedding without considering their intrinsic relationship, especially for the widely used RoPE method.

\section{Additional Experiment}
\label{Additional Experiment}

\begin{figure*}
    \centering
    \begin{subfigure}[b]{0.5\linewidth}
        \includegraphics[width=\linewidth]{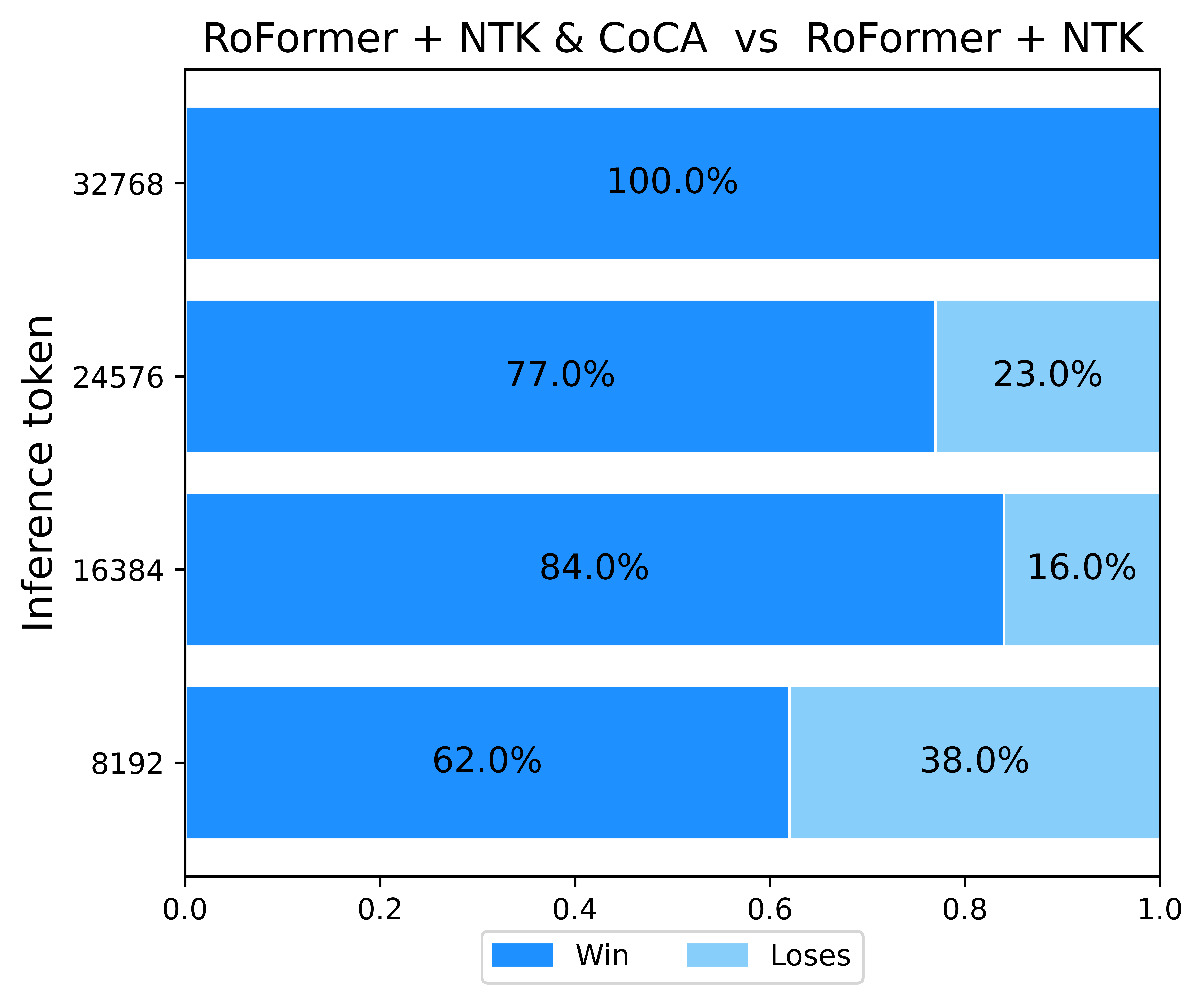}
        \caption*{(a) Inserting passkey inside 512 tokens away from end tokens}
        \label{fig:passkey_inside_512}
    \end{subfigure}%
    \begin{subfigure}[b]{0.5\linewidth}
        \includegraphics[width=\linewidth]{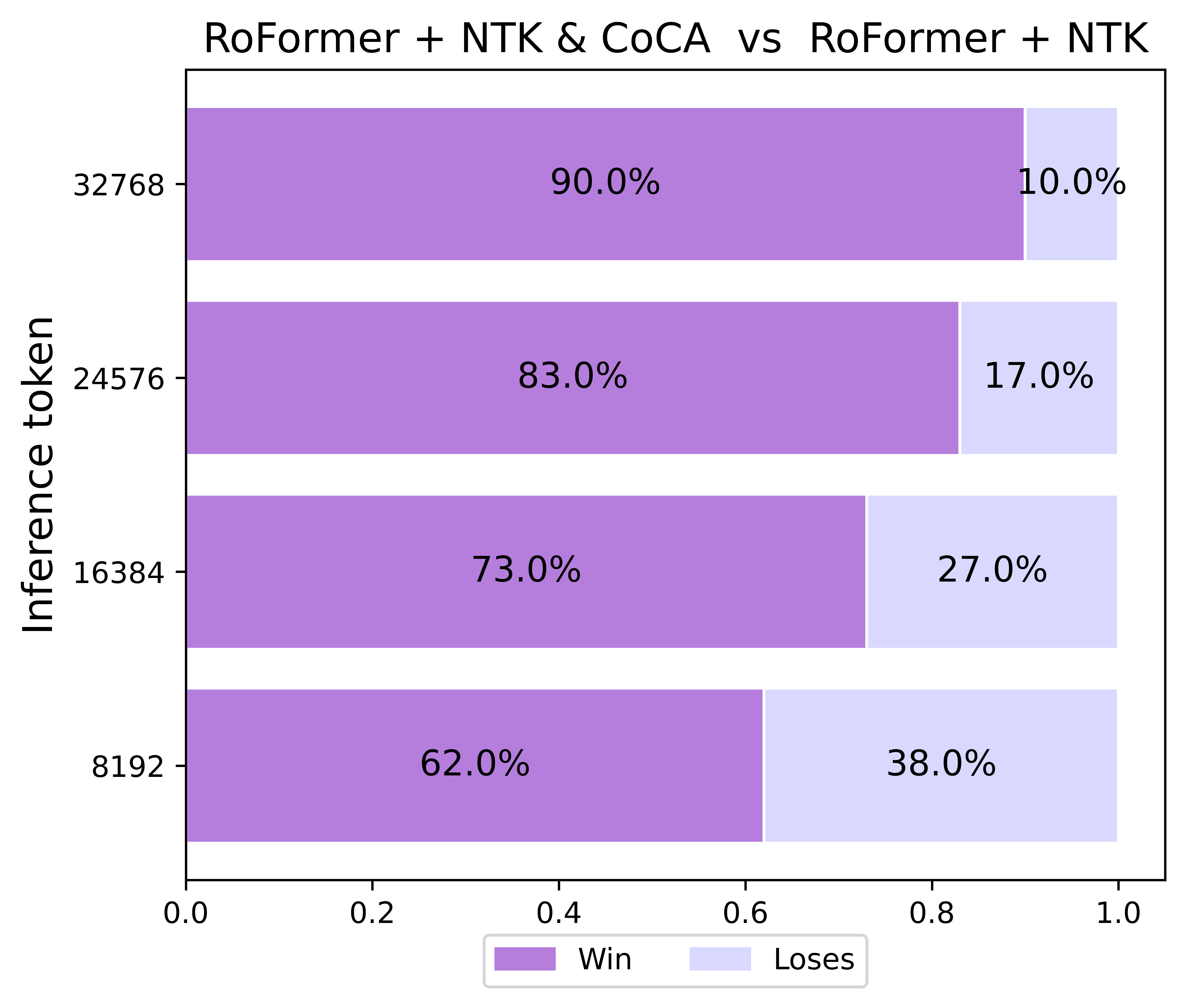}
        \caption*{(b) Inserting passkey outside 512 tokens away from end tokens}
        \label{fig:passkey_outside_512}
    \end{subfigure}
    \caption{Comparison of effective context window between RoFormer + NTK and RoFormer + NTK \& CoCA.}
    \label{fig:passkey_position}
\end{figure*}

\begin{figure}
    \centering
    \includegraphics[width=0.8\linewidth]{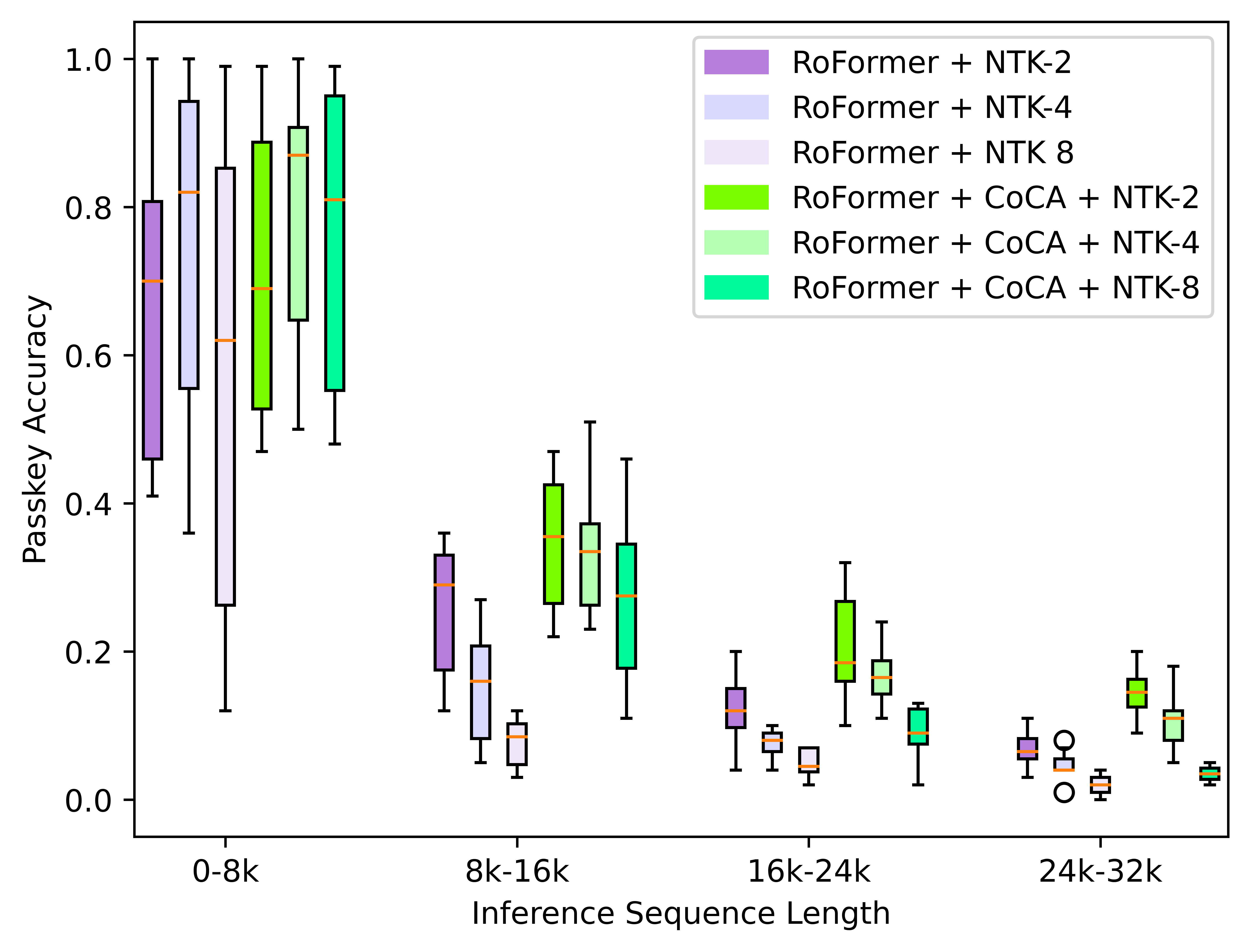}
    \caption{Passkey accuracy distribution on 4 range of distances. CoCA outperforms RoFormer for all distances and scaling factors of NTK.}
    \label{fig:passkey}
\end{figure}

\subsection{Passkey Retrieval Task Definition}
\label{passkey task}

\begin{lstlisting}[caption={Prompt format for passkey retrieval \citep{DBLP:journals/corr/abs-2305-16300}. The passkey is randomly generated from 10,000 to 99,999.},label={lst:format}]
There is an important info hidden inside a lot of irrelevant text. Find it and memorize them. I will quiz you about the important information there.

The grass is green. The sky is blue. The sun is yellow. Here we go. There and back again.
@
\vdots \quad \textcolor{commentgray}{// Repeat x times.}
@

@
\textcolor{commentgray}{// Passkey is 5 randomly generated numbers.}
@ 
The passkey is @\sethlcolor{aigold}\hl{12345}@. Remeber it. @\sethlcolor{aigold}\hl{12345}@ is the passkey.

The grass is green. The sky is blue. The sun is yellow. Here we go. There and back again. 
@
\vdots \quad \textcolor{commentgray}{// Repeat y times.}
@

What is the passkey?
\end{lstlisting}

The passkey retrieval task, as proposed by \citet{DBLP:journals/corr/abs-2305-16300}, involves the model recovering a randomly generated passkey hidden in a long document (see Listing \ref{lst:format} for the task prompt format). 
Given a language model, we can determine the effective context window by assessing the upper and lower bounds. 
We assume a random passkey is $k$ tokens away from the end of the input. 
If a model consistently fails to recover the passkey in multiple attempts, it suggests a context window size smaller than $k$. 
Conversely, successful retrievals indicate an effective context window size of at least $k$ tokens \citep{Chen2023ExtendingCW}.

In our experiments, we generate test samples based on the prompt template in Listing \ref{lst:format}, with lengths ranging from 512 to 32k. There are 100 test cases for each length. Given a language model, we input the passkey task prompt, examine the model's output for the new 64 tokens, and calculate the accuracy.

\subsection{Analysis I : Consistency of Optimization in Position Embedding}

The passkey retrieval results are presented in Section \ref{Long Context Retrieval}. Our model demonstrates superior passkey retrieval accuracy compared to baseline models under various conditions. However, we remain intrigued about its optimization, specifically whether it occurs within or beyond the confines of the training context window. To probe this further, we categorize the experimental data into two segments: passkey distance shorter and farther than the training context window length.

Figure \ref{fig:passkey_position} (a) illustrates the comparison results when the passkey is inserted less than 512 tokens away from the end token, while Figure \ref{fig:passkey_position} (b) illustrates that outside this range. When the passkey is inserted outside the 512 window, RoFormer+NTK \& CoCA consistently outperforms Roformer+NTK across various lengths of inference sequences. This superiority persists when the passkey is inserted inside the 512 window. Notably, with an increase in the length of the inference sequence, RoFormer + NTK \& CoCA demonstrates increasingly superior performance compared to RoFormer + NTK. These experiments suggest that our model can consistently optimize the position embedding and extend the effective context window.

\subsection{Analysis II : Impact of Dynamic-NTK in CoCA}
We utilize the dynamic NTK method \citep{dynamicNTK} during the inference process, applying it separately to both our model and the baseline model. To comprehensively assess the robustness of these models, we conduct a thorough validation by varying scaling factors (2, 4, and 8).

The results in Figures \ref{fig:ppl} and \ref{fig:passkey} demonstrate that, with the integration of the dynamic NTK method, our model achieves higher passkey accuracy and lower perplexity. Additionally, when the scaling factor varies between 2, 4, and 8, the vanilla RoFormer model fails to maintain stable performance. 
In contrast, CoCA consistently outperforms RoFormer at different scaling rates. This consistent trend indicates that our model is more robust, showing minimal performance fluctuations with changes in the scaling factor.

Furthermore, it suggests that by implementing collinear constraints, we can cleverly address anomalous behavior in RoPE, allowing RoPE to better leverage other extrapolation techniques.

\subsection{Analysis III : Compatibility of CoCA with PI}
\label{fine-tune PI}

\subsubsection{Experiment Setup}

We conduct experiments utilizing the pre-trained LLaMA-7B model \citep{Touvron2023LLaMAOA} and LLaMA-7B + CoCA from Section \ref{Model Variants}. To apply PI , we follow the settings of \citet{Chen2023ExtendingCW}: We set the fine-tuning sequence length to 32,768. The learning rate is adjusted to $2e-5$ with no decay to match. All other settings are maintained as the LLaMA-7B configuration. All experiments are conducted with 32 A100 GPUs, setting a per-device batch size to 1 without gradient accumulation. The experiments take 6,000 steps to accomplish.

\subsubsection{Long Context Validation}

\begin{table*}[t]
  \centering
   \small
   \setstretch{1.25}
  \begin{tabular*}{\textwidth}{@{\extracolsep{\fill}}ccccccccc}
    \toprule
    \multicolumn{2}{c}{\textbf{Method}} &
      \multicolumn{1}{l}{\textbf{512}} &
      \multicolumn{1}{l}{\textbf{1024}} &
      \multicolumn{1}{l}{\textbf{2048}} &
      \multicolumn{1}{l}{\textbf{4096}} &
      \multicolumn{1}{l}{\textbf{8192}} &
      \multicolumn{1}{l}{\textbf{16384}} &
      \multicolumn{1}{l}{\textbf{32768}} \\ \hline
    \multicolumn{9}{c}{\textit{Performance on Long Sequence Modeling (Perplexity)}} \\ \hline
    &
      LLaMA-7B+PI & 9.06 &
      7.55 &
      7.74 &
      7.16 &
      7.04 &
      6.93 &
      7.11 \\
    \multirow{-2}{*}{} &
      \quad + CoCA \& PI &
      9.65 &
      8.19 &
      8.37 &
      7.87 &
      7.84 &
      7.83 &
      7.96 \\ \hline
    \multicolumn{9}{c}{\textit{Performance on Long Context Retrieval (Passkey Accuracy)}} \\ \hline
    &
      LLaMA-7B+PI &
      1.0 &
      1.0 &
      1.0 &
      1.0 &
      1.0 &
      1.0 &
      0.99 \\
    \multirow{-2}{*}{} &
      \quad + CoCA \& PI &
      1.0 &
      1.0 &
      1.0 &
      1.0 &
      1.0 &
      0.99 &
      0.99 \\ \bottomrule
  \end{tabular*}
  \caption{Comparison results for LLaMA-7B+PI and LLaMA-7B+CoCA \& PI after fine-tuning with sequence length of 32,768. CoCA succeeds in maintaining the performance of PI within fine-tuning window size.}
  \label{tab:llama PI}
\end{table*}

The results of fine-tuning with PI are presented in Table \ref{tab:llama PI}. 
In terms of long sequence modeling, both LLaMA-7B+PI and LLaMA-7B+CoCA \& PI demonstrate competitive performance across sequence lengths ranging from 512 to 8192. However, at longer sequence lengths (16384 and 32768), LLaMA-7B+CoCA \& PI exhibits a slight performance advantage over LLaMA-7B+PI. For long context retrieval, both methods achieve exceptionally high accuracy, with scores approaching the ideal value of 1.0 across all sequence lengths.

Overall, these findings suggest that the integration of PI and the CoCA module with the LLaMA-7B model yields robust performance in both long sequence modeling and long context retrieval tasks. Additionally, the CoCA module demonstrates the ability to maintain performance levels comparable to PI, particularly evident at longer sequence lengths.

\subsubsection{Short Context Validation}
\label{Expression Ability Validation}
\begin{table}[t]
\small
\setstretch{1.25}
\centering
\begin{adjustbox}{width=\columnwidth}
\begin{tabular}{lccccc}
\toprule
\multicolumn{1}{c}{\textbf{Method}} & \textbf{Reasoning} & \textbf{Understanding} & \textbf{Language} & \textbf{Examination} & \textbf{Average} \\ 
\hline
LLaMA-7B                            & \textbf{48.25}     & 47.57                  & 46.41             & \textbf{29.63}       & 42.97            \\
+ CoCA                              & 45.55              & 51.14                  & 55.27             & 25.14                & 44.28            \\
+ PI                                & 44.98              & 51.54                  & 54.79             & 27.03                & 44.59            \\
+ CoCA \& PI                        & 46.88              & \textbf{51.82}         & \textbf{55.56}    & 25.31                & \textbf{44.89}   \\ 
\bottomrule
\end{tabular}
\end{adjustbox}
\caption{OpenCompass results of LLaMA-7B and its variants. Models integrated with CoCA achieved comparable performance to LLaMA-7B, leading no harm to the expression ability of the model.}
\label{tb:opencompass}
\end{table}
In addition to enhancing long-context extrapolation, it is imperative to consider the practicality and scalability of CoCA in short contexts. Hence, we evaluate our model on OpenCompass \citep{2023opencompass}, which comprises various dimensions, including reasoning, understanding, language, and examination. The results are presented in Table \ref{tb:opencompass}. 

The table demonstrates that LLaMA-7B models integrated with CoCA achieve performance comparable to the baseline LLaMA-7B across all evaluated dimensions. Specifically, the integration of CoCA yields no significant degradation in the expression ability of the model. This suggests that CoCA is effective not only in long-context scenarios but also in short-context tasks, demonstrating its versatility and suitability for practical applications.

\section{Computational and Spatial Complexity Analysis}
\label{Computational and spatial complexity}

\begin{table}[ht!]
\centering
\small
\setstretch{1.35}
\begin{adjustbox}{width=\columnwidth}
\begin{tabular}{c|cc|cc}
\toprule
\multirow{2}{*}{\textbf{Module}} & \multicolumn{2}{c|}{\textbf{vanilla self-attention}} & \multicolumn{2}{c}{\textbf{CoCA}} \\ \cline{2-5} 
                                 & \multicolumn{1}{c|}{Computational}     & Spatial     & Computational      & Spatial      \\ \hline
$\mathbf{W}_{QK(T)V}$           & \multicolumn{1}{c|}{$3Nd^2h$} & $Nd$   & \multicolumn{1}{c|}{$3Nd^2h$}             & $Nd$                          \\
T half       & \multicolumn{1}{c|}{—}                   & —       & \multicolumn{1}{c|}{{\color[HTML]{3166FF} $Ndh$}}   & {\color[HTML]{3166FF} $Nd$}   \\
T Relu       & \multicolumn{1}{c|}{—}                   & —       & \multicolumn{1}{c|}{{\color[HTML]{3166FF} $Ndh$}}   & {\color[HTML]{3166FF} $Nd$}   \\
QK(T) rotation & \multicolumn{1}{c|}{$2Ndh$}             & $Nd$   & \multicolumn{1}{c|}{$2Ndh$}                         & $Nd$                          \\
$\text{K}_{rot}=\text{Q}\circ \text{T}_{rot}$          & \multicolumn{1}{c|}{—}                   & —       & \multicolumn{1}{c|}{{\color[HTML]{3166FF} $N^2dh$}} & {\color[HTML]{3166FF} $N^2d$} \\
$\text{Q}_{rot}\text{K}_{rot}^\text{T}$          & \multicolumn{1}{c|}{$N^2dh$}            & $N^2$ & \multicolumn{1}{c|}{$N^2dh$}                        & $N^2$                        \\
Mask         & \multicolumn{1}{c|}{$N^2$}              & $N^2$ & \multicolumn{1}{c|}{$N^2$}                          & $N^2$                        \\
Softmax      & \multicolumn{1}{c|}{$N^2$}              & $N^2$ & \multicolumn{1}{c|}{$N^2$}                          & $N^2$                        \\ \bottomrule
\end{tabular}
\end{adjustbox}
\caption{The comparison of computational and spatial complexity between vanilla self-attention block and CoCA. Here, $N$ represents the sequence length, $h$ denotes the number of heads, and $d$ signifies the dimension of each head.}
\label{tb:2}
\end{table}

\begin{figure}[htp]
    \centering
    \includegraphics[width=0.8\linewidth]{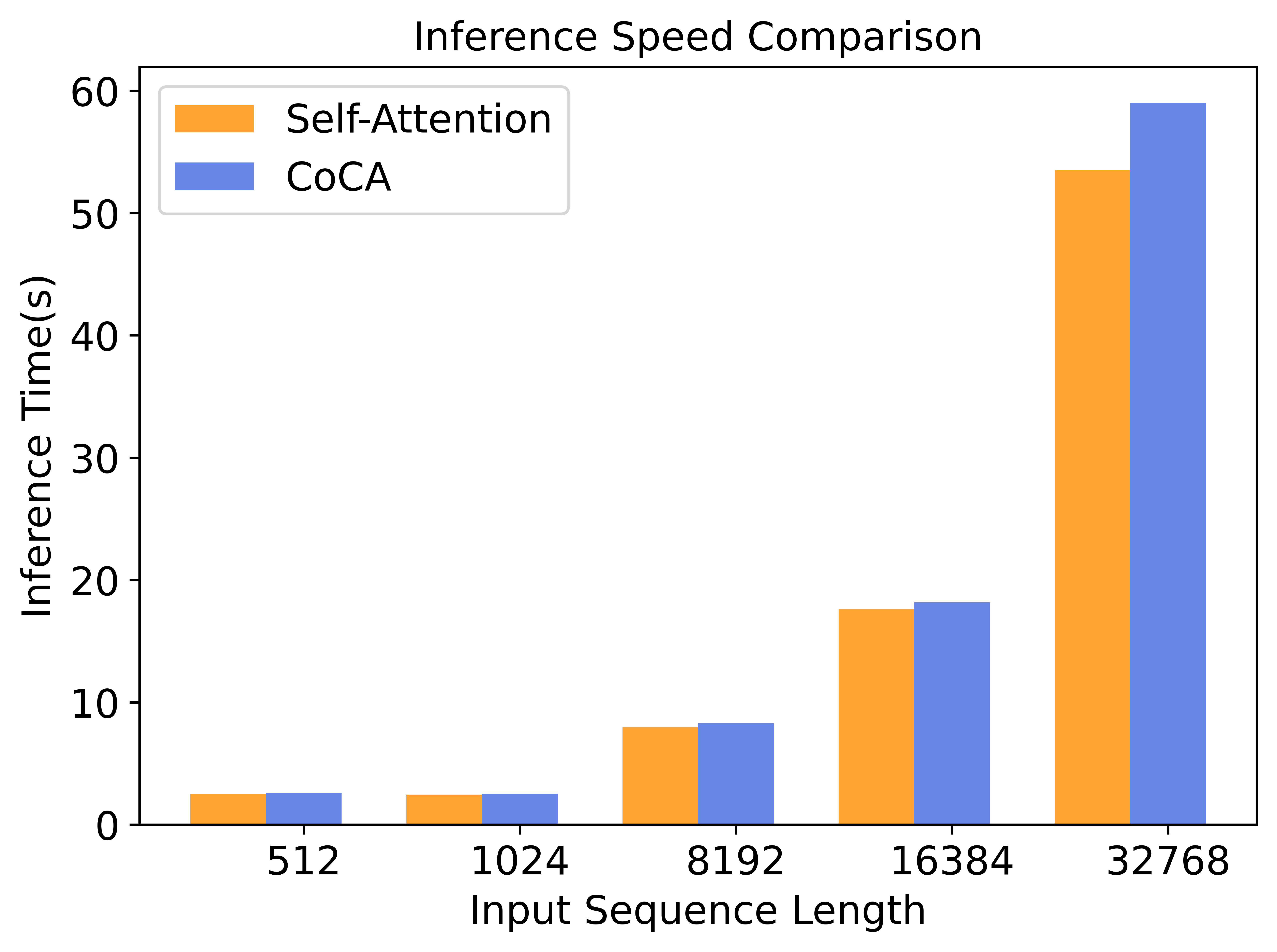}
    \caption{Inference speed comparison between CoCA and vanilla self-attention.}
    \label{fig:speed}
\end{figure}

In this section, we analyze the computational and spatial complexities of CoCA. Table \ref{tb:2} provides a detailed comparison between the vanilla self-attention mechanism and CoCA. 

When using the operation $\text{K}_{rot}=\text{Q}\circ \text{T}_{rot}$, the computational complexity of CoCA does not exceed twice that of the vanilla self-attention. 
In practice, the training and inference speed of CoCA are comparable to the vanilla self-attention mechanism, with only a slight increase of about 5\% to 10\% , as depicted in Figure \ref{fig:speed}. 
However, there is a significant increase in spatial complexity when expanding $\text{K}_{rot}=\text{Q}\circ \text{T}_{rot}$, becoming $d$ times that of the vanilla self-attention. This level of spatial complexity is not practical for applications.

To address this problem, we can draw inspiration from the computation of  $\text{Q}_{rot}\text{K}_{rot}^\text{T}$, which involves two steps: element-wise multiplication between $\text{Q}_{rot}$ and $\text{K}_{rot}$ followed by summation along the hidden dimension. Optimization is attainable by condensing the hidden dimension before fully expanding the sequence length dimension. 
Consequently, the spatial complexity is effectively reduced from $N^2d$ to $N^2$. This optimization strategy is equally applicable to $\text{K}_{rot}=\text{Q}\circ \text{T}_{rot}$. These two components can be unified as articulated in Equation (\ref{eq:15}):

\begin{equation}
\small
\begin{aligned}
\text{Q}_{rot}\text{K}_{rot}^\text{T} = \text{Q}_{rot}(\text{Q}\circ \text{T}_{rot})^\text{T}
\label{eq:15}
\end{aligned}
\end{equation}

The commendable work accomplished by opt\_einsum \citep{Smith2018} facilitates the optimization of Equation (\ref{eq:15}). Experimental results indicate that Roformer+CoCA only demands approximately 60GB of GPU memory during inference with a sequence length of 32k, aligning closely with the memory consumption of the vanilla self-attention mechanism.

\section{Theoretical Proof}

\subsection{Strong Form of Long-term Decay with CoCA}
We have introduced the basic theory of Rotary Position Embedding in Section \ref{Rotary Position Embedding}. In fact, \citep{Su2021RoFormerET} shows that RoPE has the characteristic of long-term decay:
\begin{equation}
\small
\begin{aligned}
|a(s)| &=\Bigg|\text{Re}\left[ \sum_{j=0}^{d/2-1}h_je^{is\theta_j}\right]\Bigg| \\
     &\leq (\max_i|h_{i+1}-h_i|)\sum_{j=0}^{d/2-1}|S_{j+1}|
\end{aligned}
\end{equation}
where $h_j:=(q_{2j}+iq_{2j+1})(k_{2j}-ik_{2j+1})$ and $S_j:=\sum_{k=0}^{j-1}e^{is\theta_k}$, $s=(m-n)$, $m$ for the index of query, $n$ for the index of key. Since the value of $\sum_{j=0}^{d/2-1}|S_{j+1}|$ decays with the relative distance $s$, the attention score decays either.

This characteristic ensures the stability of RoPE during extrapolation to some extent by preventing outliers. For CoCA, a stronger deduction can be formulated as follows:
\begin{equation}
\small
\begin{aligned}
|a(s)| &\leq (\max_i|l_{i+1}-l_i|)\sum_{j=0}^{d/2-1}|C_{j+1}|
\label{eq:10}
\end{aligned}
\end{equation}
where $l_j:=|q_{2j}+iq_{2j+1}||k_{2j}+ik_{2j+1}|$, and $C_j:=\sum_{k=0}^{j-1}\text{cos}(s\theta_k)$. Furthermore, it holds that:
\begin{equation}
\small
\begin{aligned}
|l_{i+1}-l_i| &\leq |h_{i+1}-h_i|
\label{eq:11}
\end{aligned}
\end{equation}

\noindent \textit{\textbf{Proof}}: 
Notice that when the initial angle $\Theta_j$ between $\mathbf{q}_j$ and $\mathbf{k}_j$ is $0$, from Equation (\ref{eq:magnitude dependence}), the attention score can be simplified as:
\begin{equation}
\small
\begin{aligned}
a(s) &=\text{Re}\left[ \sum_{j=0}^{d/2-1}h_je^{is\theta_j}\right] \\
     &=\sum_{j=0}^{d/2-1}l_j\cos(s\theta_j)
\end{aligned}
\end{equation}

By following the study of \citep{Su2021RoFormerET}, we can easily derive the estimation in Equation (\ref{eq:10}).

For Equation (\ref{eq:11}), applying the triangle inequality, we get:
\begin{equation}
\small
\begin{aligned}
|h_{i+1}-h_i|\geq||h_{i+1}|-|h_i||
\end{aligned}
\end{equation}
Reviewing the definition of $h_i =(q_{2j}+iq_{2j+1})(k_{2j}-ik_{2j+1})$, we will find:
\begin{equation}
\small
\begin{aligned}
|h_{i+1}-h_i|   &\geq||h_{i+1}|-|h_i|| \\
                &=||\mathbf{q}_{i+1}\mathbf{k}^*_{i+1}|-|\mathbf{q}_{i}\mathbf{k}^*_{i}|| \\
                &=||\mathbf{q}_{i+1}\mathbf{k}_{i+1}|-|\mathbf{q}_{i}\mathbf{k}_{i}|| \\
                &=|l_{i+1}-l_i|
\end{aligned}
\end{equation}

\subsection{Rotary Borders Analysis}
\label{Rotary Borders Analysis}
\begin{figure}[htp]
    \centering
    \includegraphics[width=0.7\linewidth]{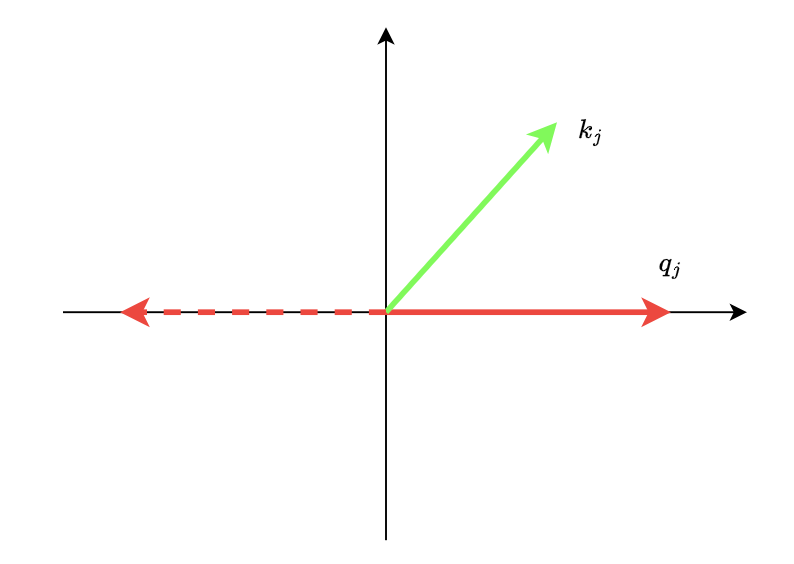}
    \caption{Rotary Borders Analysis. Regarding $\mathbf{q}_j$ as $x$-axis, 3 distinct boundaries correspond to $\mathbf{k}_j$, $-\mathbf{q}_j$, and $\mathbf{q}_j$}
    \label{fig:6}
\end{figure}
In Section \ref{anomalous analysis}, we analyzed the anomalous phenomena of RoPE. To illustrate the rotation anomalies, let's focus on a specific instance (case (d) of Section \ref{anomalous analysis}). As shown in Figure \ref{fig:6}, three distinct boundaries emerge during the rotation. By adopting a relative coordinate system with $\mathbf{q}_j$ serving as the $x$-axis, these boundaries correspond to $\mathbf{k}_j$, $-\mathbf{q}_j$, and $\mathbf{q}_j$. 

Everytime the relative angle of $\mathbf{q}_j$ and $\mathbf{k}_j$ crosses these boundaries, the monotonicity of their inner-product $<\mathbf{q}_j,\mathbf{k}_j>$ undergoes a reversal. Thus, for the vanilla self-attention, it learnt a piecewise monotonic function of $<\mathbf{q}_j,\mathbf{k}_j>$:

\begin{equation}
\small
<\mathbf{q}_j,\mathbf{k}_j>=\left\{
\begin{aligned}
&\uparrow(m-n), \forall -(2\pi-\Theta_j)\leq\theta{(\mathbf{q}_j,\mathbf{k}_j)}<0 \\
&\downarrow(m-n), \forall 0\leq\theta{(\mathbf{q}_j,\mathbf{k}_j)}<\pi \\
&\uparrow(m-n), \forall \pi \leq\theta{(\mathbf{q}_j,\mathbf{k}_j)}<2\pi \\
&... \\
&\uparrow(m-n), \forall (2k-1)\pi\leq\theta{(\mathbf{q}_j,\mathbf{k}_j)}<(2k)\pi \\
&\downarrow(m-n), \forall (2k)\pi\leq\theta{(\mathbf{q}_j,\mathbf{k}_j)}<(2k+1)\pi \\
\end{aligned}
\right.
\label{eq:monotonic}
\end{equation}
where $\theta{(\mathbf{q}_j,\mathbf{k}_j)} = \Theta_j + (m-n)\theta_j$ defined in Section \ref{anomalous analysis}.

This introduces confusion into the model during direct context extrapolation. Therefore, methods like PI and NTK tried to introduce interpolation or extrapolation techniques to eliminate out-of-distribution (OOD) positions.

Except the first equation in Equation (\ref{eq:monotonic}), the two boundaries caused by $-\mathbf{q}_j$, and $\mathbf{q}_j$ are regular with periodicity of $2\pi$, it is easy to handle when applying methods like PI or NTK. However, the boundaries caused by $\mathbf{k}_j$ are hard to handle. There are $d/2*h*L$ ($d$ for head dimension, $h$ for number of heads, $L$ for number of layers) different boundaries during context extrapolation, which break the periodicity of $2\pi$. 

Furthermore, after applying interpolation or extrapolation techniques, more positions will fall into this abnormal area. It increased $k$ times ($k$ for interpolation factor) for PI and $\lambda^{2j/d}$ times ($\lambda$ for scaling factor) for NTK. 

From this perspective, positional concentration of PI resulted in more trouble than NTK, i.e. additionally more positions in abnormal area during context extrapolation. This may explain in some extent why NTK could be used without fine-tuning for vanilla self-attention, but PI requires fine-tuning.

By enforcing $\Theta_j$ to $0$, our proposed CoCA, constraining $\mathbf{k}_j$ to be collinear with $\mathbf{q}_j$, effectively resolves the border-related challenge associated with $\mathbf{k}_j$.

From experiments in Secton \ref{Experiment Results}, with the integrating of CoCA, now NTK can be leveraged well through direct use, while PI achieved improvement for direct use but still limited, which requires further studies.


\subsection{Homeomorphism of Representation Space}
\label{Homeomorphism of representation space}

\begin{theorem}
    (Homeomorphism of representation space) For any attention score defined as follows:
    \begin{equation}
    \small
        \begin{aligned}
        a(m,n) &=\text{\rm{Re}}(\langle f(\mathbf{q}_m,m), f(\mathbf{q}_m,n) \circ \mathbf{t}_n \rangle)
        \label{Homeomorphism 1}
        \end{aligned}
    \end{equation}
    where $\mathbf{q}_m$ is the query, $m$ is the index number of query, $\mathbf{t}_n$ is the collinear coefficient of CoCA, $n$ is the index number of key, $f$ is the rotation operator.

    Denote its representation space with respect to $\mathbf{q}_m$ as:

    \begin{equation}
    \small
        \begin{aligned}
        F(Q) = \{a(m,n)|\forall \mathbf{q}_m \in Q \subset \mathbb{R}^d\}
        \label{Homeomorphism 2}
        \end{aligned}
    \end{equation}

    where $\mathbf{q}_m={\textbf{W}}_Q\mathbf{x}_m$, $\mathbf{x}_m \in \mathbb{E}_N$, $m \in [1,N]$ and $\mathbb{E}_N$ is the word embedding space, ${\textbf{W}}_Q$ is the projection matrix.

    Then we have the following homeomorphism:
    \begin{equation}
    \small
        \begin{aligned}
        F(Q) \cong F(Q_{half})
        \label{Homeomorphism 3}
        \end{aligned}
    \end{equation}

    where $Q_{half}=Q|_{q_{2j}=q_{2j+1},\forall j \in [0,d/2-1]}$.

\label{Homeomorphism}
\end{theorem}

\noindent \textit{\textbf{Proof}}: 
We prove it by demonstrating the homeomorphism mapping $\mathcal{G}$:

\begin{equation}
\small
\begin{aligned}
\mathcal{G} : F(Q) &\rightarrow F(Q_{half}) \\
F((q_0,...,q_{d-1})
 &\mapsto F((\sqrt{\frac{q_0^2+q_1^2}{2}},...,\sqrt{\frac{q_{d-2}^2+q_{d-1}^2}{2}}) \\
\end{aligned}
\label{Homeomorphism 4}
\end{equation}

It consists of three parts:

\noindent \textit{Part I} ($\mathcal{G}$ is a bijection): recall Equation (\ref{eq:magnitude dependence}), we have:
\begin{equation}
\small
\begin{aligned}
\mathcal{G}(X)=X,\forall X \in F(Q)
\end{aligned}
\label{Homeomorphism 5}
\end{equation}
which implies that $\mathcal{G}$ is an identity mapping, naturally injective.

Next, we prove that $\mathcal{G}$ is also surjective: for any $Y=F((q_0,...,q_{d-1})|_{q_{2j}=q_{2j+1}}) \in F(Q_{half})$, there exists $\widetilde{Y} \in F(Q)$ such that $\mathcal{G}(\widetilde{Y})=Y$. Let
\begin{equation}
\small
\begin{aligned}
\widetilde{Y}=F((q_0,...,q_{d-1})|_{q_{2j}=q_{2j+1}}) \in F(Q)
\end{aligned}
\label{Homeomorphism 6}
\end{equation}
obviously we have $\mathcal{G}(\widetilde{Y})=Y$.

\noindent \textit{Part II} ($\mathcal{G}$ is continuous): For any $X_0 \in F(Q)$, $\epsilon>0$, there exists $\delta$, such that if $|X-X_0|<\delta$, then $|\mathcal{G}(X)-\mathcal{G}(X_0)|<\epsilon$.

From \textit{Part I}, $\mathcal{G}$ is an identity mapping, let $\delta=\epsilon$, then the continuity of $\mathcal{G}$ holds.

\noindent \textit{Part III} ($\mathcal{G}^{-1}$ is continuous): $\mathcal{G}$ is an identity mapping, so is $\mathcal{G}^{-1}$. Following Part II, we immediately deduce that $\mathcal{G}^{-1}$ is continuous. \qedsymbol

\subsection{Slack Position Embedding}
\label{Slack Position Embedding}

Let $\mathcal{H}$ be a Hilbert space, and $\{\mathcal{T}(n)|n \geq 0\}\subset \mathcal{L}(\mathcal{H})$ is a family of bounded linear operator on $\mathcal{H}$. $\mathcal{A}$ is the inner-product defined on $\mathcal{H}$.

If it satisfies the following property, then we call $\{\mathcal{T}(n)|n \geq 0\}$ is a relative (bounded linear) operator on $\mathcal{H}$:
\begin{equation}
\small
\begin{aligned}
&\begin{aligned} \\
\exists \ \{\mathcal{S}(m)|m\in \mathbb{Z}\} : \mathcal{H} \times \mathcal{H} &\rightarrow \mathbb{C} \\
(X,Y) &\mapsto \mathcal{S}(m)(X,Y) \\
&\end{aligned} \\
& \text{is a family of semi-bilinear operator on } \mathcal{H} \\
&\begin{aligned} \\
&s.t. \ \mathcal{S}(p-q)(X,Y) =  \mathcal{A}(\mathcal{T}(p)(X), \mathcal{T}(q)(Y)) \\
&\forall \ p,q \in [0,N],  X,Y \in \mathcal{H}, \\
&\end{aligned} \\
\end{aligned}
\label{eq:spe1}
\end{equation}
Additionally, if it satisfies the following property, then we call $\{\mathcal{T}(n)|n \geq 0\}$ is a slack relative (bounded linear) operator on $\mathcal{H}$:
\begin{equation}
\small
\begin{aligned}
&\begin{aligned} \\
\exists \ \{\mathcal{S}(m)|m\in \mathbb{Z}\} : \mathcal{H} \times \mathcal{H} &\rightarrow \mathbb{C} \\
(X,Y) &\mapsto \mathcal{S}(m)(X,Y) \\
&\end{aligned} \\
& \text{is a family of semi-bilinear operator on } \mathcal{H} \\
&\begin{aligned} \\
&and \ \mathcal{H}' \subset \mathcal{H},\mathcal{H}'\neq \varnothing  \\
&s.t. \ \mathcal{S}(p-q)(X,Y) =  \mathcal{A}(\mathcal{T}(p)(X), \mathcal{T}(q)(Y)) \\
&\forall \ p,q \in [0,N],  X,Y \in \mathcal{H}', \\
&\end{aligned} \\
\end{aligned}
\label{eq:spe2}
\end{equation}
Specifically, when $\mathcal{H}$ represents our projection space in self-attention, and $\{\mathcal{T}(n)|n \geq 0\}$ is a position embedding on it, such as the Rotary Position Embedding (RoPE), we refer to it as a Slack Position Embedding (SPE) if it satisfies the property described in Equation (\ref{eq:spe2}).

\end{document}